%% file: main.tex
\documentclass{article}

\usepackage{microtype}
\usepackage{graphicx}
\usepackage{subfigure}
\usepackage{booktabs} 

\usepackage{hyperref}

\usepackage[accepted]{icml2023}
\usepackage{amsmath}
\usepackage{amssymb}
\usepackage{mathtools}
\usepackage{amsthm}

\usepackage[capitalize,noabbrev]{cleveref}

\theoremstyle{plain}

\theoremstyle{definition}

\theoremstyle{remark}

\usepackage{enumitem}
\usepackage{graphicx}
\usepackage{amsmath}
\usepackage{amssymb}
\usepackage{array}
\usepackage{setspace}
\usepackage{subfigure}
\usepackage{multirow}
\usepackage{tikz}
\usepackage{rotating}
\usepackage{wrapfig}
\usepackage{xcolor}
\usepackage{pifont}
\newcommand*\circled[1]{\tikz[baseline=(char.base)]{\node[shape=circle,fill=black,text=white,draw,inner sep=.1pt] (char) {#1};}}
\usepackage{listings}

\usepackage[textsize=tiny]{todonotes}

\icmltitlerunning{Prioritized Trajectory Replay: A Replay Memory for Data-driven Reinforcement Learning}

\begin{document}

\twocolumn[
\icmltitle{Prioritized Trajectory Replay: A Replay Memory for \\Data-driven Reinforcement Learning}



\begin{icmlauthorlist}
\icmlauthor{Jinyi Liu\textsuperscript{\rm +}}{yyy,fuxi}
\icmlauthor{Yi Ma}{yyy}
\icmlauthor{Jianye Hao}{yyy}
\icmlauthor{Yujing Hu}{fuxi}
\icmlauthor{Yan Zheng}{yyy}
\icmlauthor{Tangjie Lv}{fuxi}
\icmlauthor{Changjie Fan}{fuxi}
\textsuperscript{\rm +}{\texttt{jyliu@tju.edu.cn}}
\end{icmlauthorlist}

\icmlaffiliation{yyy}{College of Intelligence and Computing, Tianjin University}
\icmlaffiliation{fuxi}{NetEase Fuxi AI Lab}

\icmlcorrespondingauthor{Jianye Hao}{jianye.hao@tju.edu.cn}

\icmlkeywords{Machine Learning, ICML}

\vskip 0.3in
]



\printAffiliationsAndNotice{}  

\begin{abstract}
In recent years, data-driven reinforcement learning (RL), also known as offline RL, have gained significant attention. However, the role of data sampling techniques in offline RL has been overlooked despite its potential to enhance online RL performance. Recent research suggests applying sampling techniques directly to state-transitions does not consistently improve performance in offline RL. Therefore, in this study, we propose a memory technique, (Prioritized) Trajectory Replay (TR/PTR), which extends the sampling perspective to trajectories for more comprehensive information extraction from limited data. TR enhances learning efficiency by backward sampling of trajectories that optimizes the use of subsequent state information. Building on TR, we build the weighted critic target to avoid sampling unseen actions in offline training, and Prioritized Trajectory Replay (PTR) that enables more efficient trajectory sampling, prioritized by various trajectory priority metrics. We demonstrate the benefits of integrating TR and PTR with existing offline RL algorithms on D4RL. In summary, our research emphasizes the significance of trajectory-based data sampling techniques in enhancing the efficiency and performance of offline RL algorithms.
\end{abstract}
\section{Introduction}

Reinforcement learning~(RL) has made significant developments in recent years and has been widely applied in various scenarios~\cite{YeLSSZWYYWGCYZS20,KiranSTMSYP22,SalvatoFMP21}. 
From a functional perspective, researchers usually study efficient RL algorithms from three aspects~\cite{HesselMHSODHPAS18}: data collection, data sampling, and training algorithms.
Data collection encompasses various methods for obtaining or generating diverse and comprehensive data, such as exploration~\cite{exploraton2023jianye} and data augmentation~\cite{LaskinLSPAS20}.
Data sampling refers to studying different sampling schemes based on existing data, to improve the learning efficiency~\cite{SchaulQAS15PER, LeeCC19EBU}. Training algorithms are what most researchers focus on, aiming to maximize the optimization objectives through various techniques, including $Q$-value estimation~\cite{MnihKSRVBGRFOPB15}, representation~\cite{SangTMHZMLW22}, and more.

To overcome the high cost of interacting with the environment in real scenarios, offline RL has received wide attention as a way to learn good policy based on the fixed dataset. The study of offline RL can also be classified into the aforementioned three categories, but with slight differences. In offline RL, the literature mainly focuses on training algorithms through conservative estimation~\cite{CQLKumarZTL20, Bai0YDG0W22}, network structure~\cite{DTChenLRLGLASM21}, and other aspects. In terms of data collection, learning only from fixed dataset does not require exploration for new data, and research in data augmentation receives certain attention~\cite{wang2022bootstrapped}. 
Nevertheless, research on data sampling techniques under offline settings is still in its early stages and lacks a unified and comprehensive conclusion~\cite{fu2022benchmarking}. 
Experience from online RL reveals that data sampling techniques can improve performance~\cite{SchaulQAS15PER}. Therefore, this study primarily focuses on data sampling techniques to gain a more comprehensive understanding of their role in offline RL.

\begin{figure*}[t]
  \centering
  \subfigure{\includegraphics[height=99px]{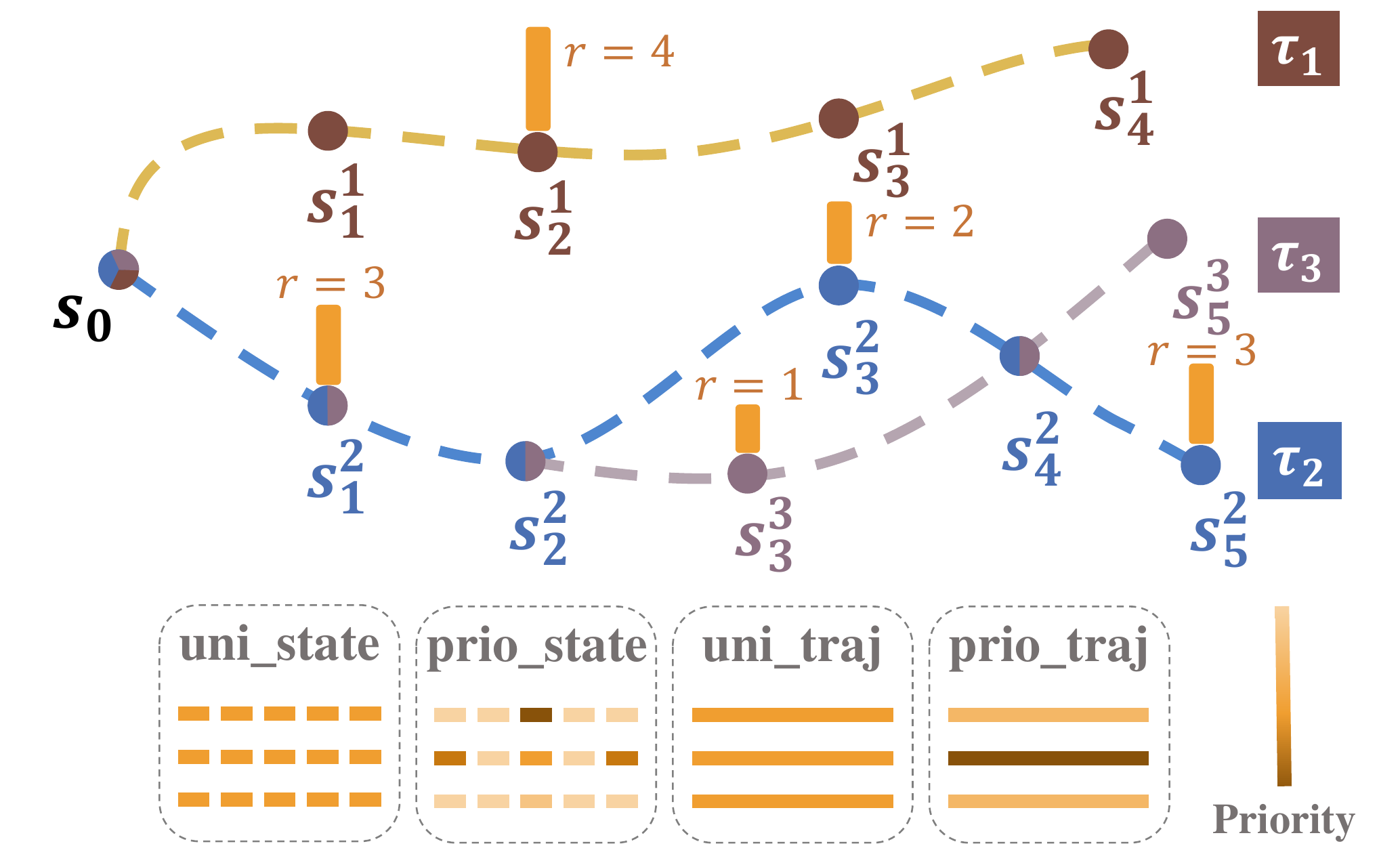}} 
  \subfigure{\includegraphics[height=99px]{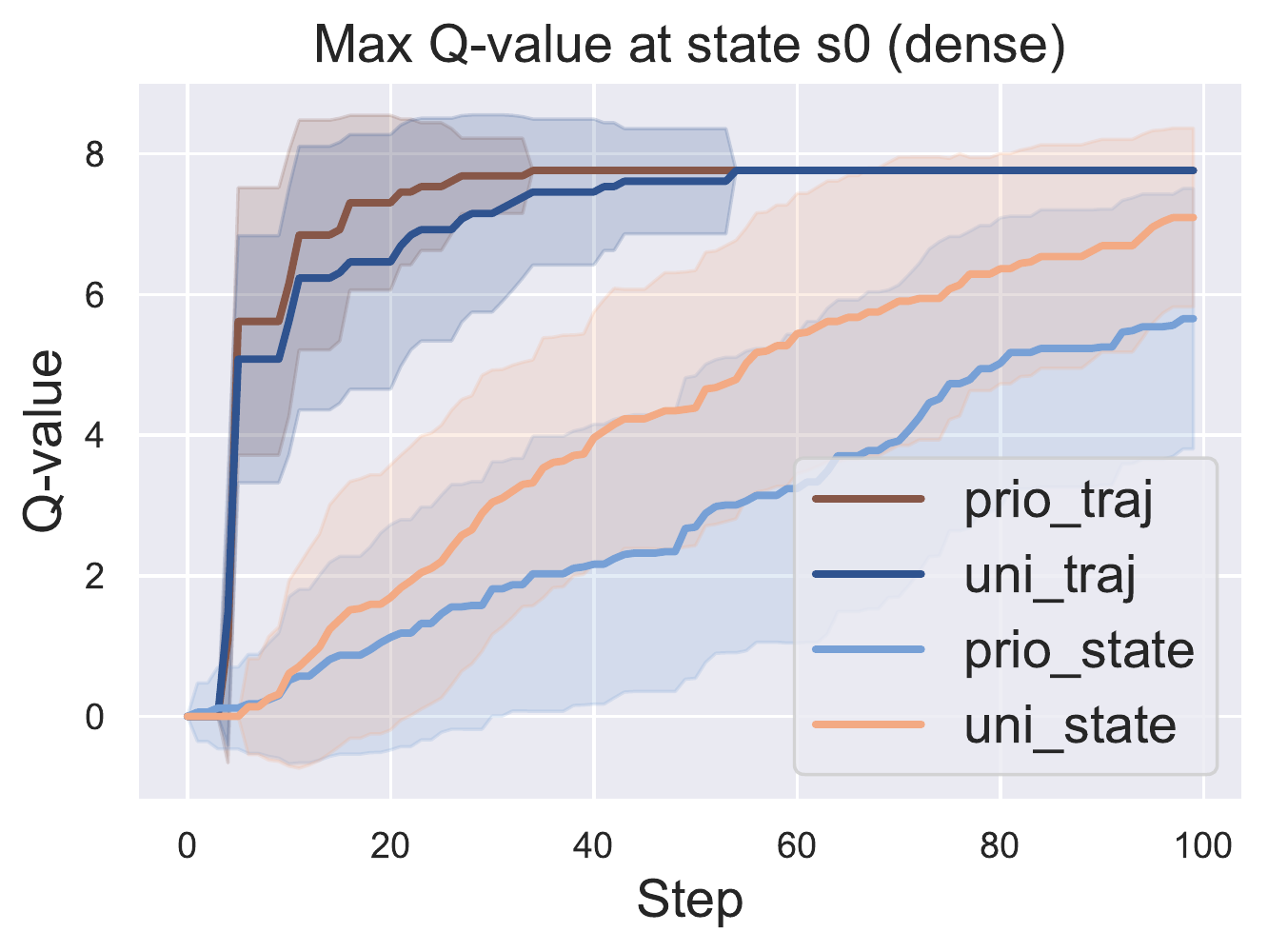}} 
  \subfigure{\includegraphics[height=99px]{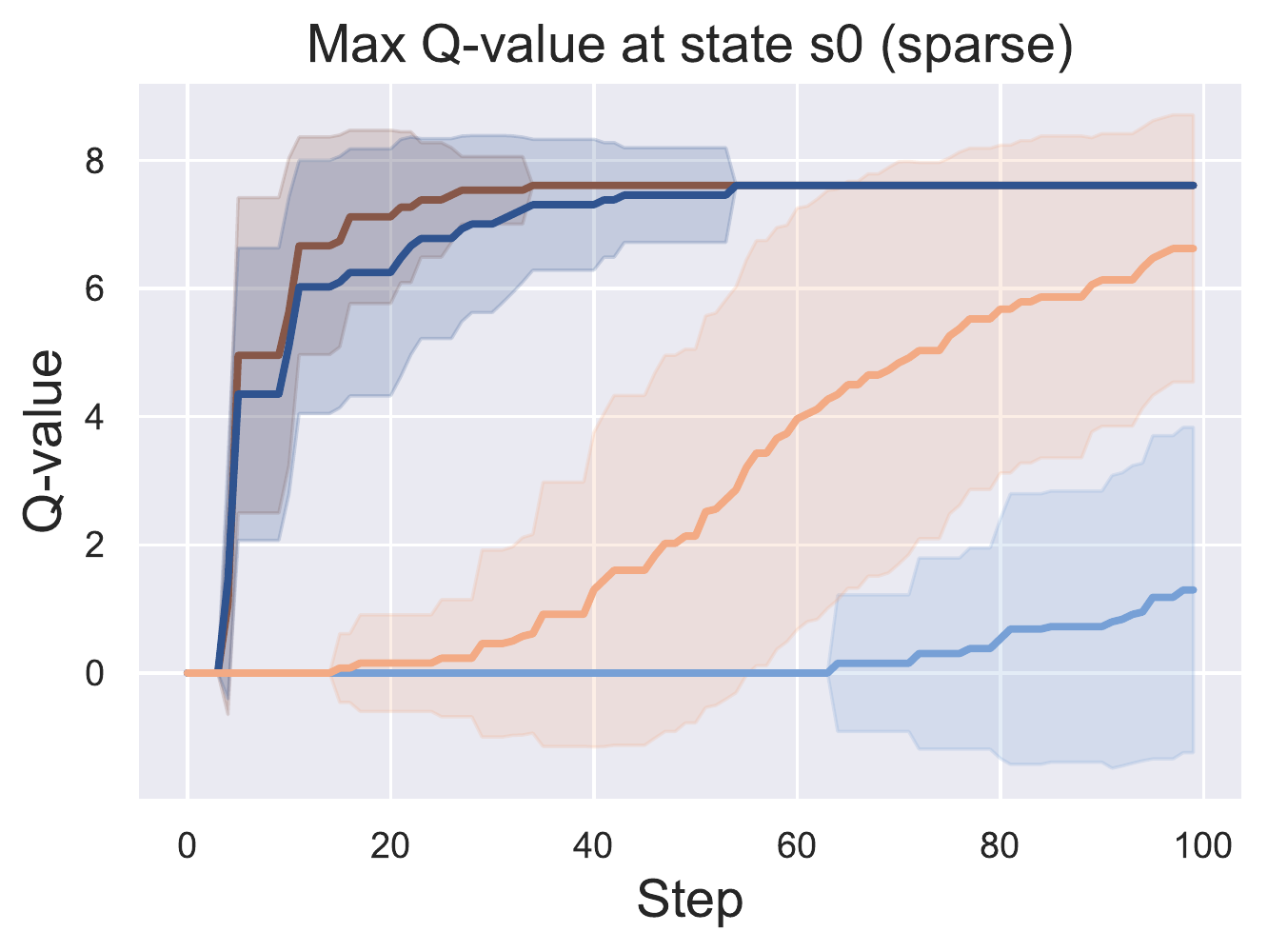}}
  \caption{A motivating example on finite data. \textbf{Left}: The illustration of state transitions in three trajectories $\tau_i$ started at state $s_0$, with the reward $r$ for labeled state or 0 for others, and four different sampling techniques. \textbf{Middle and right}: Curves of the estimated maximum $Q$-value at state $s_0$ learned on these three trajectories. The solid line is the averaged value over 50 seeds, and the shaded area the standard variance. The oracle value, taking into account the discount factor, is slightly less than 8.}
  \label{fig:fig1}
\vskip -0.2in
\end{figure*}

We first present an illustrative example in Figure~\ref{fig:fig1} to analyze the effect of data sampling techniques on limited data. The results indicate that sampling state-transitions directly, whether being uniform (\textit{uni\_state}) or prioritized (\textit{prio\_state}), results in inferior learning of the $Q$-value at the initial state $s_0$. Even prioritized sampling (\textit{prio\_state}) offers little advantage over uniform sampling, which aligns with the conclusion of~\cite{fu2022benchmarking}. Such sampling could potentially overlook critical states (such as $s_1$) that link other states to the initial state $s_0$, leading to ineffective propagation of subsequent reward signals to $s_0$. In contrast, when sampling along trajectories (\textit{uni\_traj} and \textit{prio\_traj}), the propagation of reward signals to $s_0$ is accelerated, converging faster to the optimal value. Such advantage is more prominent on sparse reward trajectories, where reward is given only at the last state. Compared to uniform trajectory sampling~(\textit{uni\_traj}), \textit{prio\_traj} efficiently samples the trajectory with higher returns, further enhancing training efficiency. 
Our observation suggest that it is vital to consider the entire trajectory when determining the order of data sampling. {{Such trajectory perspective provides more informative insights into offline RL, rather than solely focusing on individual $(s, a)$ pairs.}}

To fully discover the potential of data sampling techniques from trajectory perspective for offline RL algorithms, we propose a plug-and-play memory, Prioritized Trajectory Replay (PTR), and conduct a detailed experimental analysis on these techniques, ultimately improving the performance of existing offline RL algorithms.
We begin by implementing Trajectory Replay (TR), the basic memory to store and sample data in trajectories. Specifically, we sample and evaluate the states within a trajectory in a backward order. For instance, given two consecutive states, the latter one is sampled and evaluated first, followed by the former one. This allows each state to benefit from the estimated subsequent states, leading to improved convergence speed and performance especially on tasks with sparse rewards.

Based on TR, we continue to explore the potential of data sampling from two perspectives. From the algorithm updating perspective, we are inspired by EBU~\cite{LeeCC19EBU} and modify the target $Q$ computation method by balancing the original target $Q$-value and SARSA $Q$ target value~\cite{sutton2018reinforcement}. The SARSA $Q$-value is calculated to limit the update process to use only trajectory data, avoiding the sampling of out-of-distribution (OOD) actions, thus alleviates the extrapolation error in offline learning. From the trajectory sampling perspective, we further propose Prioritized Trajectory Replay (PTR), considering the characteristics of trajectories (e.g., quality or uncertainty) and verifies different trajectory priority definitions. 
Through sufficient evaluation, we discover that prioritizing the sampling of trajectories with higher upper quartile mean or mean value of rewards, or trajectories with lower uncertainty, is effective in accelerating the learning of offline RL algorithms.

The main contribution of this paper is a concrete analysis of several trajectory-based data sampling techniques in offline RL, which are integrated into a unified memory structure - PTR. We emphasize that this work does not aim to propose a universally state-of-the-art (SOTA) algorithm. Rather, our study examines the advantage of basic trajectory sampling in TR, and the role of various trajectory priorities in PTR. We hope that our findings can provide useful insights towards developing powerful data sampling techniques in the future.

\section{Related Work}

This section provides an overview of fundamental data sampling methods and offline RL algorithms, which inspire our considerations of trajectory-based priorities in offline RL. 

\paragraph{Data sampling in online RL} Data sampling studies how to efficiently sample data to improve training. In RL, this can be traced back to~\cite{MooreA93}, which improves planning efficiency by selecting the next updated state based on priorities. For model-free RL, PER~\cite{SchaulQAS15PER} proposes to replay transitions with high temporal-difference (TD) error more frequently, which significantly improves the training efficiency of DQN~\cite{MnihKSRVBGRFOPB15}. Based on PER, several algorithms have been proposed, which optimize the error-based sampling~\cite{PanMFWYR022, abs-2102-03261}, or propose other priority metrics~\cite{OhGSL18, fu2022benchmarking}, etc. These algorithms mostly focus on priority-based state-transition sampling. Other approaches study data sampling methods from the perspective of trajectory, and the most typical ones are EBU~\cite{LeeCC19EBU} and TER~\cite{Hong0LPA22TER}, which propose backward sampling from the end to the beginning of the trajectory. Backward sampling can make the update process more timely and accurately utilize the reward signals of the subsequent states to achieve accelerated learning. 

\paragraph{Offline RL} In recent years, a large amount of research has focused on offline RL algorithms that learn promising policies from a fixed offline dataset. A key issue that offline RL algorithms need to address is how to avoid the impact of estimation errors (extrapolation errors) of unseen data in the dataset on policy learning, since collecting new data is impossible~\cite{offlinelevine, FujimotoMP19}. Common solutions can be divided into several categories: the first category imposes a policy constraint, which constrains the distribution of learned policies to be similar to the distribution of behavioral policies in the dataset~\cite{KumarFSTL19, TD3BCFujimotoG21}; the second category applies support constraint, which require the action values of the learned policy to be restricted to the actions that appear in the dataset~\cite{IQLKostrikovNL22, xiao2023the}; the third category involves conservative $Q$-value~\cite{CQLKumarZTL20, LiTTZ21, EDACAnMKS21, Bai0YDG0W22} or MDP model~\cite{COMBOYuKRRLF21} estimation, which avoids selecting OOD actions during the policy update process by conservatively estimating unseen data.

In offline training, most research focuses on designing training algorithms, with less attention given to data sampling techniques. 
The literature~\cite{fu2022benchmarking} evaluates several data sampling techniques from a traditional state-transition perspective based on TD3+BC, but consistent conclusions have not been obtained. Considering the return of trajectories, ReD~\cite{Boostingyue} uses return-based data re-balance for to improve the sampling probability of transitions on trajectories with higher returns. Determining the sampling order based on the trajectory return can be regarded as a special case of our study. Under the perspective of trajectory, we approach data sampling from two aspects: backward sampling on trajectories, and prioritized trajectory sampling based on trajectory quality or uncertainty.

\section{Preliminary}
\subsection{Offline Reinforcement Learning}

Offline RL refers to RL conducted on offline data. In offline RL, we have collected several trajectories, called the dataset $\mathcal{D}=\{\tau_j\}_{j=1}^N$. Each trajectory is a sequence of state-action-reward with length $l$: $\tau=\{(s_i, a_i, r_i)\}_{i=0}^{l-1}$, where $s_i$ represents the state at time $i$, $a_i$ represents the action taken by the agent at that state, $r_i$ represents the immediate reward, and $s_{i+1}$ represents the next state.
The optimization objective of offline reinforcement learning is usually to maximize the expected cumulative rewards through the dataset $\mathcal{D}$, without collecting new data. 

\subsection{Prioritized Experience Replay}

Off-policy RL algorithms store historical experience in a replay memory. In traditional training, training data is uniformly sampled from the replay memory~\cite{MnihKSRVBGRFOPB15}. PER (Prioritized Experience Replay)~\cite{SchaulQAS15PER} prioritizes sampling data from the memory based on the TD-error of state-transitions, leading to significant improvement in DQN training efficiency. Specifically, the distribution $P(j)$ used for sampling transition $j$ is defined as follows:
\begin{equation}
\begin{aligned}
P(j) = &\frac{p_j^\alpha}{\sum_k p_k^\alpha}, s.t., p_k = |r_k + \gamma_k Q_\mathrm{target}(s_{k+1}, \\&\arg\max_a Q(s_{k+1}, a)) - Q(s_{k}, a_{k})| + \epsilon
\end{aligned}
\end{equation}
where $p_j$ is the priority of transition $j$, $\alpha$ determines how much prioritization is used, $\gamma$ is the discount factor, $Q(\cdot)$ and $Q_\mathrm{target}(\cdot)$ represent the $Q$-value and target $Q$-value for state-action pair $(s, a)$, and $\epsilon$ is a small constant that prevents priorities from being zero. In this way, PER can prioritize high-priority transitions based on the scale of TD-error, thereby accelerating the agent's learning process.

\input{sec/method}
\input{sec/exp}

\input{sec/con}

\bibliography{main}
\bibliographystyle{icml2023}

\newpage
\appendix
\input{sec/appendix}


\end{document}

%% file: sec/method.tex
\section{Trajectory Replay}
\vskip 0.2in
\begin{figure*}[t]
  \centering
  \includegraphics[width=0.75\textwidth]{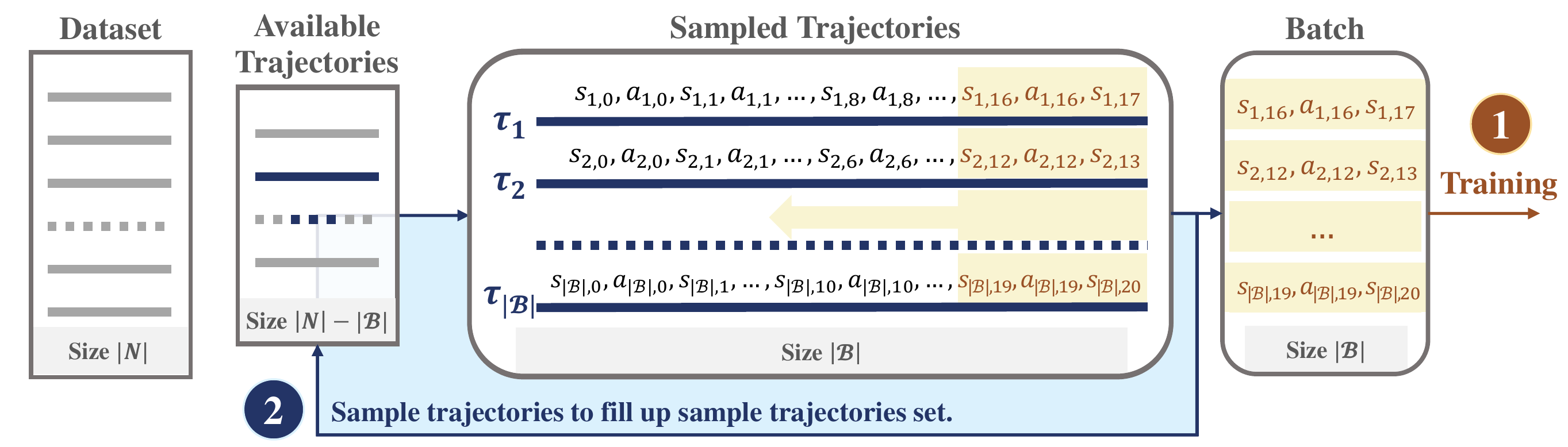}
  \vskip -0.04in
  \caption{Overview of the process of data sampling based on Trajectory Replay.}
  \label{fig:TR-based-backward}
  \vskip -0.2in
\end{figure*}

In this section, we introduce Trajectory Replay (TR), a replay memory for storing offline data as complete trajectories and sampling batch data from a trajectory perspective. First, we describe the details of TR, including the technique of backward trajectory sampling used to sample from stored trajectories. We then introduce a weighted target computation based on TR, akin to EBU~\cite{LeeCC19EBU}.

\subsection{Trajectory Replay}
To implement batch data sampling based on trajectory sequence, we maintain a basic replay memory storing the data in the form of trajectory, called Trajectory Replay, which differs from the traditional memory storing data as separate transitions. 
In EBU~\cite{LeeCC19EBU}, it is suggested that sampling data in backward order on a trajectory is beneficial for online learning. Therefore, in this paper, we adopt backward sampling along this insight, as the default sampling order of TR. Such sampling based on TR is expected to offer certain advantages by making more timely and comprehensive use of information from successor state transitions.

Figure~\ref{fig:TR-based-backward} illustrates the process of data sampling based on TR, in which $|N|$ trajectories in the offline dataset are stored. The available trajectories set stores all trajectories that have not been sampled. At the beginning, we sample $|\mathcal{B}|$  trajectories from it and add them to the sampled trajectories set. At each time step, the last state transition of $|\mathcal{B}|$ trajectories are moved to the data batch $\mathcal{B}$ for algorithm training. 
When all transitions of a trajectory in the sampled trajectories set have been added to $\mathcal{B}$, a new trajectory is required to be sampled from the available trajectories set to to fill up the sampled trajectories set.
In such basic TR-based data sampling, the critic update process is not changed. Given the data batch $\mathcal{B}$, a typical optimization objective following deterministic actor-critic algorithm~\cite{SilverLHDWR14} is as follows:
\begin{equation}
\begin{aligned}
    &\mathcal{J}(\theta) = \mathbb{E}_{(s_t, a_t, r_t, s_{t+1})\sim\mathcal{B}}\left[\ \frac{1}{2} \left(Q_\theta(s_t, a_t) - Q_{\mathrm{tg}}(s_t, a_t)\right)^2\ \right], \\
    &Q_{\mathrm{tg}}(s_t, a_t) = r_{t} + \gamma Q_{\Bar{\theta}}(s_{t+1}, \pi(s_{t+1})),
\end{aligned}
\label{eq:q_target}
\end{equation}
where $\theta$ and $\Bar{\theta}$ represent the parameters of critic network and target critic network, respectively, and $\pi(\cdot)$ represents the policy distribution.

Trajectory replay is the foundation for studying data sampling techniques from a trajectory perspective in this study. As shown in Figure~\ref{fig:TR-based-backward}, in the following, we mainly study \circled{1}~the training process based on TR in Section~\ref{sec:weighted_target}, and \circled{2}~the process of sampling trajectories according to different trajectory priority metrics in Section~\ref{sec:ptr}.

\subsection{Weighted Backward Update based on TR}
\label{sec:weighted_target}
A key challenge in offline RL is the extrapolation error arising from $Q$-value estimation. Although TR implements a basic trajectory-based sampling process, it does not take into account alleviating the extrapolation error, which can limit offline RL performance. Our research shows that a SARSA-like target can be introduced based on TR's trajectory backward sampling process to completely avoid accessing OOD actions when calculating TD error during training. The critic target in Eq.~\ref{eq:q_target} is now as follows:
\begin{equation}
\begin{aligned}
    Q_{\mathrm{target}}(s_t, a_t) &= r_{t} + \gamma Q_{\mathrm{target}}(s_{t+1}, a_{t+1}), 
\end{aligned}
\label{eq:q_target_sarsa}
\end{equation}
where $Q_{\mathrm{target}}(s_{t+1}, a_{t+1})$ is the target $Q$-value, calculated in the next time step in the same trajectory. Optimizing Eq.~\ref{eq:q_target_sarsa} can be considered as an implicit support constraint as OOD actions are avoided.


Based on it, we modify the computation of the target $Q$ by balancing the original target in Eq.~\ref{eq:q_target} and the vanilla SARSA target, to introduce the implicit support constraint in Eq.~\ref{eq:q_target_sarsa}. We obtain the following weighted form of target estimation: 
\begin{equation}
\begin{aligned}
    Q_{\mathrm{target}}(s_t, a_t) = r_{t} + \gamma&[ (1-\beta) Q_{\mathrm{target}}(s_{t+1}, a_{t+1}) \\&+ \beta Q_{\Bar{\theta}}(s_{t+1}, \pi(s_{t+1}))], 
\end{aligned}
\label{eq:weighted_target}
\end{equation}
where $\beta$ is the hyper-parameter that controls how much the original target should be used.

\section{Prioritized Trajectory Replay}
\label{sec:ptr}
Building upon Trajectory Replay, we have developed a weighted target to improve learning efficiency. In this section, we shift our focus from the updating process to enhancing the trajectory sampling process depicted in Figure ~\ref{fig:TR-based-backward}. To enhance the performance through more properly sampling trajectories, we introduce several novel \textbf{trajectory priority} metrics to prioritize trajectories during sampling. These metrics primarily consider two key factors: the quality of the trajectory return and the degree of trajectory uncertainty.

Instead of uniformly sampling from the available trajectories, we introduce a probabilistic sampling approach that incorporates trajectory priority metrics for trajectory $\tau_j$, denoted as $\mathrm{pri}(\tau_j)$, to prioritize trajectories. 
To avoid potential biases towards high priority trajectories, we use the ranking order, denoted as $\mathrm{rank}(\mathrm{pri}(\tau_j))$, as a measure to determine the sampling probability $p_{\tau_j}$, rather than relying solely on the absolute value of $\mathrm{pri}(\tau_j)$. Assigning appropriate priority to each trajectory through its rank ensures that all available trajectories are able to be utilized effectively. To this end, we define the prioritized sampling probability distribution $P(\tau_j)$ for all available trajectories as follows:
\begin{equation}
\begin{aligned}
P(\tau_j) &= \frac{p_{\tau_j}^\alpha}{\sum_k p_{\tau_k}^\alpha},\ \text{s.t.},\ p_{\tau_j} = \frac{1}{\mathrm{rank}(\mathrm{pri}(\tau_j))}. \\
\end{aligned}
\end{equation}

\subsection{Priority based on Trajectory Quality}
Higher quality trajectories have been found to be more beneficial for offline RL~\cite{FujimotoMP19}. To this end, we propose utilizing trajectory quality as a priority metric during the data sampling process to prioritize high-quality trajectories. 
Recognizing that using return as a metric may unfairly prioritize longer trajectories, we introduce alternative measures such as the mean of rewards on the trajectory, including upper quartile mean~(UQM) and upper half mean~(UHM). 
In this regard, we define six distinct metrics to assess trajectory quality, as follows:

\begin{itemize}[leftmargin=0.5cm]
    \item \textbf{Return}. The undiscounted sum rewards of the trajectory. 
    \item \textbf{Avg. reward}. The averaged reward of the trajectory.
    \item \textbf{UQM reward}. The averaged reward of the top 25\% state transitions in the trajectory.
    \item \textbf{UHM reward}. The averaged reward of the top 50\% state transitions in the trajectory. 
    \item \textbf{Min reward}. The minimum reward of the trajectory.  
    \item \textbf{Max reward}. The maximum reward of the trajectory.
\end{itemize}
Trajectory priority $\mathrm{pri}(\tau_j)$ is established based on these metrics to prioritize the sampling of high-quality trajectories. In particular, \textit{Return} and \textit{Avg. reward} reflect the overall performance of the trajectory, while \textit{UQM reward}, \textit{UHM reward}, and \textit{Max reward} reflect the quality of the best transitions in the trajectory. on the other hand, \textit{Min reward} prioritizes trajectories with higher minimum reward values, indicating a more stringent criterion for high quality.

\subsection{Priority based on Trajectory Uncertainty}

Within the context of offline RL, uncertainty serves as a pivotal metric by quantifying the level of disagreement present among $Q$ estimates. Such disagreement indicates a lack of confidence in the estimated $Q$-value for a given state-action pair. Thus, a higher degree of uncertainty indicates greater unreliability in the knowledge of the state-action pair, making it unwise and infeasible to use it for policy execution.
Taking a trajectory perspective, we define trajectory priority based on uncertainty, such that trajectories exhibiting lower degrees of uncertainty are assigned greater priority during sampling. Specifically, we establish trajectory priority metrics based on the mean, upper quartile mean~(UQM), and lower quartile mean~(LQM) of all state-action pairs' uncertainty values present within the trajectory, as follows:
\begin{itemize}[leftmargin=0.5cm]
    \item \textbf{Lower mean unc}. The reciprocal of the average uncertainty of all state-action pairs.
    \item \textbf{Lower LQM unc}. The reciprocal of the average uncertainty of bottom 25\% state-action pairs.
    \item \textbf{Lower UQM unc}. The reciprocal of the average uncertainty of top 25\% state-action pairs.
\end{itemize}
Alternatively, we can assign higher priority to trajectories with higher levels of uncertainty. We can obtain the following trajectory priority metrics by taking the reciprocals of the aforementioned values: \textit{Higher mean unc.}, \textit{Higher LQM unc.}, and \textit{Higher UQM unc.}, respectively.

By incorporating trajectory priority, we upgrade TR to \textbf{Prioritized Trajectory Replay~(PTR)}, achieving probabilistic sampling of trajectories.
We anticipate that prioritizing trajectories with higher quality would be more suitable for sparse reward tasks, where the reward signals  exhibit significant variations on these trajectories and thus require an emphasis on trajectories with higher value for learning during probabilistic sampling. 
Also, we expect that prioritizing trajectories with lower uncertainty can be effective in dense reward tasks, which can aid in accurately estimating the level of uncertainty. Conversely, prioritizing trajectories with higher uncertainty is not expected to work well, as errors from data with high uncertainty can accumulate severely for finite data and hinder learning.  Section~\ref{sec:exp} provides a comprehensive and detailed empirical evaluation of these distinct trajectory priority metrics.


%% file: sec/exp.tex
\section{Experiment}
\label{sec:exp}
To reveal the consistency between our analysis and the performance of TR/PTR, and demonstrate the advantage over other advanced methods, we conduct experiments to address the following questions:

\begin{itemize}[topsep=0pt, partopsep=0pt, leftmargin=*]
    \item[] \textbf{\textit{RQ1} (Trajectory replay)}: Can basic sampling of TR bring an improvement to offline RL algorithms by efficiently using information from successor state transitions?
    \item[] \textbf{\textit{RQ2} (Weighted target based on TR)}: Can modifying the critic target to SARSA-style target or weighted target bring further improvement to the performance, and how do they demonstrate?
    \item[] \textbf{\textit{RQ3} (Prioritized trajectory replay)}: Without changing the critic target, can the prioritized trajectory sampling bring consistent performance improvement, and what are the characteristics of different trajectory priority metrics?
\end{itemize}

\subsection{Baseline Algorithms and Implementation Details}
\paragraph{Benchmark} In order to comprehensively analyze the advantages and disadvantages of various proposed data sampling mechanisms from a trajectory perspective, we conduct empirical evaluations on the D4RL benchmark~\cite{D4RL}, focusing primarily on the Mujoco \verb|-v2| datasets (dense reward), Adroit \verb|-v1| datasets (sparse reward), and AntMaze \verb|-v0| datasets (sparse reward). 

\paragraph{Baselines} Baseline algorithms include (i) TD3+BC, which uses regularization terms of behavior cloning to constrain the learning policy, (ii) IQL, which uses expectile regression to focus on in-sample actions and avoids querying the values of unseen actions, and (iii) EDAC, which implements conservative estimation of critic based on ensemble $Q$ networks. 
The most critical hyper-parameters in TD3+BC is $\alpha$ used to control the weights of RL and imitation learning. On Mujoco tasks, we set $\alpha$ to 2.5 and the reproduced results are similar to those reported in the original paper. On the Antmaze and Adroit datasets, we conduct a hyperparameter search in the ranges {0.0001, 0.05, 0.25, 2.5, 25, 36, 50, 100}, resulting in better results than previously reported. The most important hyper-parameters of EDAC algorithm are the $n$ used to control the number of ensembles, and $\eta$, the weight of the ensemble gradient diversity term. To make it easier to verify the advantage, we set $n$ to 3 for \verb|mujoco-medium-replay-v2| datasets and 10 for others. We use the default hyper-parameters for IQL, as recorded in the original paper. All hyper-parameters mentioned above are listed in the Appendix.

\paragraph{Implementation details} To ensure code conciseness, readability, and a fair and identical experimental evaluation across algorithms, we reproduce the baseline algorithms based on the CORL repository~\cite{tarasov2022corl} and implement various sampling techniques. Unless otherwise specified, all results reported in the following for the baseline algorithms are the results we have reproduced ourselves. For our methods, various data sampling techniques are implemented based on a plug-and-play memory module, TR/PTR. The only hyper-parameter $\beta$ is used to construct the weighted critic target, and we conduct grid searches mainly using values from {0.05, 0.25, 0.5, 0.75, 0.95, 0.98} in the experiments. In basic TR and trajectory-based priority sampling, no hyper-parameters needs to be fine-tuned. The supplementary materials include our code.

\subsection{Evaluation of Trajectory Replay}
\begin{table}[t]
  \caption{Normalized average returns of TR and baselines on Gym Mujoco, Antmaze, Adroit tasks in D4RL. The averaged performance and standard deviation of 3 runs are reported.}
\vskip 0.15in
  \label{tab:tab1}
  \renewcommand{\arraystretch}{1.0}
  \centering
  \scriptsize
\begin{tabular}{>{\raggedright}m{0.9cm}|>{\raggedleft}m{0.8cm}>{\raggedleft}m{0.8cm}|>{\raggedleft}m{0.68cm}>{\raggedleft}m{0.68cm}|>{\raggedleft}m{0.68cm}>{\raggedleft}m{0.68cm}}
\toprule
\textbf{Task Name} & \textbf{TD3+BC} & \textbf{TD3+BC\\(TR)} & \textbf{EDAC} & \textbf{EDAC\\(TR)} & \textbf{IQL} & \textbf{IQL\\(TR)}\tabularnewline
\midrule
{Mujoco} & \textbf{659.81} & 630.54 & \textbf{617.71} & 546.44 & \textbf{691.41} & 669.99 \tabularnewline
{Antmaze} & 98.36 & \textbf{223.67} & - & - & 329.90 & \textbf{356.47} \tabularnewline
{Adroit} & \textbf{540.41} & \textbf{545.95} &  2.37 & 23.41 & \textbf{545.70} & \textbf{543.52}\tabularnewline
\bottomrule
\end{tabular}
\vskip -0.1in
\end{table}

We first evaluate our proposed TR by implementing the most basic trajectory perspective-based data sampling technique with trajectory backward sampling, which serves as the basis for follow-up on critic target and priority trajectory sampling. We present the reproduced results of the baseline algorithm and the performance improvements achieved by combining TR with these algorithms in Table~\ref{tab:tab1}, and full results of all datasets in Appendix. 

The results suggest that backward sampling based on TR is particularly effective in tasks with sparse rewards. Consistent performance improvements of TR are shown in the Antmaze and Adroit datasets, with notable score increases achieved by TD3+BC, EDAC, and IQL on Antmaze of over 100, 20, and 30 points respectively. These findings support the efficacy of trajectory backward sampling as an approach for efficiently utilizing reward information from subsequent state transitions, which is consistent with previous research on online scenarios.

However, for dense reward tasks, the benefits of TR are limited. On datasets \verb|walker2d-medium-replay-v2| and \verb|hopper-medium-replay-v2|, there is even a noticeable decrease in performance. This highlights the urgent need for deeper investigate of data sampling techniques based on TR to improve offline RL algorithms' performance.

\subsection{Evaluation of Backward Weighted Update}
\begin{table*}[t]
  \caption{Normalized average returns of SARSA-style and weighted target critic based on TR on Gym Mujoco, Antmaze, Adroit tasks in D4RL. The averaged performance and standard deviation of 3 runs are reported.}
\vskip 0.15in
  \label{tab:tab2}
  \renewcommand{\arraystretch}{0.96}
  \centering
  \scriptsize
\begin{tabular}{>{\raggedright}m{3.25cm}|>{\raggedleft}m{1.65cm}>{\raggedleft}m{1.65cm}>{\raggedleft}m{1.65cm}>{\raggedleft}m{1.65cm}|>{\raggedleft}m{1.2cm}>{\raggedleft}m{1.2cm}} 
\toprule
\textbf{Task Name} & \textbf{TD3+BC}\\\textbf{(Reproduced)} & \textbf{TD3+BC}\\\textbf{(TR)} & \textbf{TD3+BC}\\\textbf{(SARSA)} &\textbf{ TD3+BC}\\\textbf{(weighted)}  & \textbf{TD3+BC}\\\textbf{(Paper)} & \textbf{IQL}\\\textbf{(Paper)} \tabularnewline 
\midrule
halfcheeh-medium-v2 & 48.20$\pm$0.37 & 48.13$\pm$0.38 & 45.85$\pm$0.30 & 48.13$\pm$0.38  & 48.3 & 47.4 \tabularnewline 
hopper-medium-v2 & 60.47$\pm$5.76 & 59.96$\pm$2.01 & 61.60$\pm$7.73 & 61.60$\pm$7.73  & 59.3 & 66.3 \tabularnewline 
walker2d-medium-v2 & 84.12$\pm$1.68 & 82.95$\pm$2.51 & 84.63$\pm$4.89 & 84.63$\pm$4.89  & 83.7 & 78.3 \tabularnewline 
halfcheetah-medium-replay-v2 & 44.76$\pm$0.31 & 43.91$\pm$0.79 & 34.95$\pm$1.71 & 43.91$\pm$0.79  & 44.6 & 44.2 \tabularnewline 
hopper-medium-replay-v2 & 50.60$\pm$23.04 & 49.83$\pm$18.01 & 31.15$\pm$9.72 & 49.83$\pm$18.08  & 60.9 & 94.7 \tabularnewline 
walker2d-medium-replay-v2 & 77.83$\pm$16.17 & 36.14$\pm$33.99 & 46.45$\pm$25.02 & 86.28$\pm$2.47  & 81.8 & 73.9 \tabularnewline 
halfcheetah-medium-expert-v2 & 89.26$\pm$4.48 & 93.63$\pm$1.04 & 52.34$\pm$4.69 & 93.63$\pm$1.04  & 90.7 & 86.7 \tabularnewline 
hopper-medium-expert-v2 & 94.01$\pm$10.39 & 105.50$\pm$6.16 & 105.50$\pm$6.16  & 105.50$\pm$6.16  & 98.0 & 91.5 \tabularnewline 
walker2d-medium-expert-v2 & 110.57$\pm$0.68 & 110.51$\pm$0.37 & 106.80$\pm$3.10 & 110.76$\pm$0.32  & 110.1 & 109.6 \tabularnewline 
\midrule
\textbf{Total} & 659.81  & 630.54 & 567.96 &\textbf{ 684.26}  & 677.4 & \textbf{692.4} \tabularnewline 
\midrule
\midrule
antmaze-umaze-v0 & 32.00$\pm$52.01 & 91.53$\pm$3.36 & 71.17$\pm$3.63 & 92.53$\pm$2.21  & 78.6 & 87.5 \tabularnewline 
antmaze-umaze-diverse-v0 & 0.00$\pm$0.00 & 0.00$\pm$0.00 & 44.10$\pm$21.78 & 58.45$\pm$7.85  & 71.4 & 62.2 \tabularnewline 
antmaze-medium-play-v0 & 35.65$\pm$37.50 & 44.20$\pm$38.70 & 0.00$\pm$0.00 & 71.08$\pm$9.20  & 10.6 & 71.2 \tabularnewline 
antmaze-medium-diverse-v0 & 17.53$\pm$27.32 & 40.40$\pm$37.49 & 0.00$\pm$0.00 & 72.33$\pm$6.81  & 3.0 & 70.0 \tabularnewline 
antmaze-large-play-v0 & 0.00$\pm$0.00 & 32.37$\pm$14.07 & 0.00$\pm$0.00 & 28.80$\pm$6.36  & 0.2 & 39.6 \tabularnewline 
antmaze-large-diverse-v0 & 0.00$\pm$0.00 & 15.17$\pm$7.72 & 0.00$\pm$0.00 & 18.87$\pm$29.75  & 0.0 & 47.5 \tabularnewline 
\midrule 
\textbf{Total} & 98.36 & 223.67 & 19.21 & \textbf{342.06}   & 193.8 & \textbf{378.0} \tabularnewline 
\midrule
\midrule
pen-cloned-v1 & 64.21$\pm$31.82 & 71.83$\pm$32.42 & 65.67$\pm$36.25 & 75.75$\pm$34.63  & - & 37.3 \tabularnewline 
hammer-cloned-v1 & 0.62$\pm$0.51 & 0.93$\pm$1.05 & 0.70$\pm$0.76 & 0.93$\pm$1.05  & - & 2.1 \tabularnewline 
door-cloned-v1 & -0.11$\pm$1.14 & -0.08$\pm$0.05 & -0.15$\pm$0.02 & -0.08$\pm$0.05  & -& 1.6 \tabularnewline 
relocate-cloned-v1 & -0.26$\pm$0.04 & -0.24$\pm$0.03 & -0.16$\pm$0.05 & -0.04$\pm$0.06  & - & -0.2 \tabularnewline 
pen-expert-v1 & 136.77$\pm$18.28 & 138.74$\pm$5.53 & 134.61$\pm$15.15 & 145.84$\pm$13.97  & - & - \tabularnewline 
hammer-expert-v1 & 128.58$\pm$0.45 & 128.60$\pm$0.38 & 127.36$\pm$0.76 & 129.18$\pm$0.66  & - & - \tabularnewline 
door-expert-v1 & 105.75$\pm$0.78 & 105.07$\pm$1.26 & 105.69$\pm$1.92 & 105.69$\pm$1.92  & - & - \tabularnewline 
relocate-expert-v1 & 104.84$\pm$3.54 & 101.11$\pm$3.19 & 104.22$\pm$2.67 & 104.22$\pm$2.67   & - & - \tabularnewline 
\midrule
\textbf{Total} & 540.41 & 545.95 & 537.92 & \textbf{562.49}   & - & - \tabularnewline 
\bottomrule
\end{tabular}
\vskip -0.1in
\end{table*}

Following TR, we proceed to introduce two target critic forms, namely a vanilla SARSA-style target (Eq.~\ref{eq:q_target_sarsa}) and a weighted target (Eq.~\ref{eq:weighted_target}), aimed at mitigating extrapolation errors and enhancing offline RL performance.  The former strictly constrains actions by utilizing all action information in the trajectory during the update process, while the latter balances the original target and vanilla SARSA-style target, thereby relaxing the action range constraint of the former. The corresponding weight values, denoted by $\beta$, are provided in the Appendix. 

Experimental results demonstrate that the vanilla SARSA-style target based on TR, exhibits stable performance and outperforms baseline algorithms on Adroit and simple Antmaze datasets. 
However, when applied to more complex tasks, the SARSA-style target exhibits unstable performance. This instability is primarily caused by the algorithm's overly simplistic and brute-force approach of avoiding out-of-distribution actions, which is inadequate for ensuring stable optimization for complex Mujoco tasks and several challenging Antmaze tasks.

Compared to basic TR and SARSA-style target, the weighted target strategy significantly improves the performance of the baseline algorithm and overcomes the issue of performance degradation in complex tasks. It achieves the best performance on nearly all datasets, and its results are comparable to, or even better than, those achieved by IQL, which is considered to be much better than TD3+BC.

Nevertheless, a notable limitation of the weighted target is its reliance on the tuning of the weight. Thus, it is imperative to consider optimizing the target form or exploring alternative data sampling techniques based on TR.

\subsection{Evaluation of Prioritized Trajectory Sampling}

\begin{table*}[t]
  \caption{Overall performance comparison of various trajectory priority metrics based on TD3+BC and IQL. The reported results are the sum of the average performance over 3 runs.}
\vskip 0.15in
  \label{tab:PTR}
  \centering
  \footnotesize
\begin{tabular}{>{\raggedright}m{0.2cm}>{\raggedright}m{1.0cm}>{\raggedleft}m{1.0cm}|>{\raggedleft}m{0.8cm}>{\raggedleft}m{0.8cm}>{\raggedleft}m{0.8cm}>{\raggedleft}m{0.8cm}>{\raggedleft}m{0.8cm}>{\raggedleft}m{0.8cm}>{\raggedleft}m{0.8cm}>{\raggedleft}m{0.8cm}>{\raggedleft}m{0.8cm}>{\raggedleft}m{0.8cm}}
\toprule
& env.& baseline & TR & Return & Avg\_r & UQM\_r &UHM\_r & Min\_r & Max\_r & L\_mean & L\_LQM & L\_UQM\tabularnewline
\midrule
\multirow{3}{*}{\rotatebox{90}{TD3BC}} & Mujoco  & 659.81 & 630.54 & 632.17 & 639.62 & 621.69 & 630.57 & 649.92 & 604.87 & \textbf{679.88} & \textbf{690.18} & \textbf{698.59}\tabularnewline
& Antmaze  & 85.18 & 223.67 & 224.36 & \textbf{252.03} & \textbf{241.63} & 232.53 & \textbf{261.83} & 95.70 & 56.40 & 72.47 & 171.53\tabularnewline
& Adroit  & 540.41 & 545.95 & \textbf{555.20} & 546.89 & 546.41 & \textbf{560.86} & \textbf{554.66} & 536.60 & 547.03 & 549.84 & 547.23\tabularnewline
\midrule
\multirow{3}{*}{\rotatebox{90}{IQL}} & Mujoco  & 691.41 & 669.99 & \textbf{710.14} & 682.26 & 648.09 & 638.14 & 604.44 & 649.38 & \textbf{709.84} & {705.61} & \textbf{709.32}\tabularnewline
& Antmaze  & 329.90 & 356.47  & 360.45 & \textbf{385.99} & \textbf{380.33} & \textbf{376.09} & 347.57 & 280.13 & 218.01 & 195.20 & 256.28\tabularnewline
& Adroit  & 543.59 & 543.52 & \textbf{550.12} &  {545.14} & 541.79 & \textbf{545.48} & 541.31 & 544.10 & \textbf{545.38} & 541.11 & 542.53\tabularnewline
\bottomrule
\end{tabular}
\vskip -0.1in
\end{table*}

\begin{figure*}[t]
\vskip 0.2in
  \centering
  \subfigure{\includegraphics[height=73.1px]{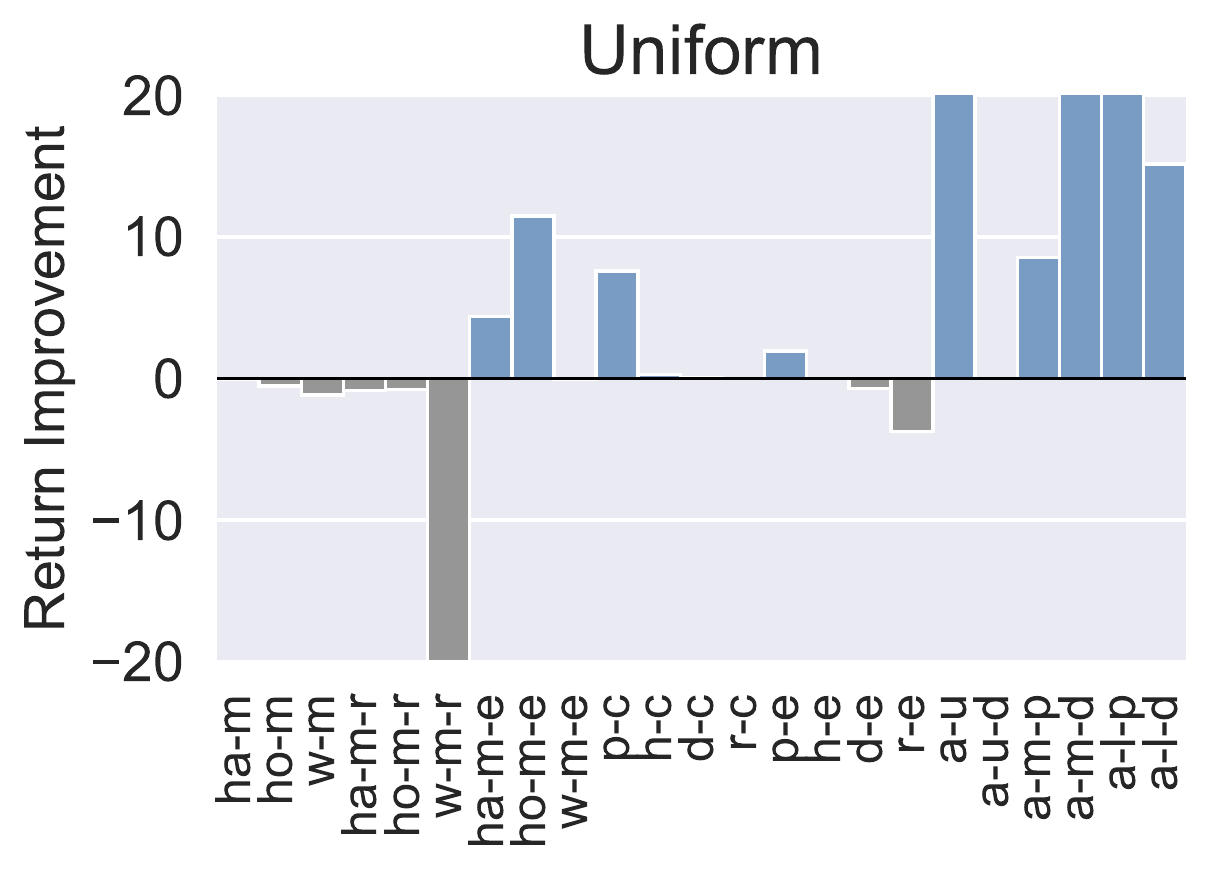}} \hspace{-6px}
  \subfigure{\includegraphics[height=73.1px]{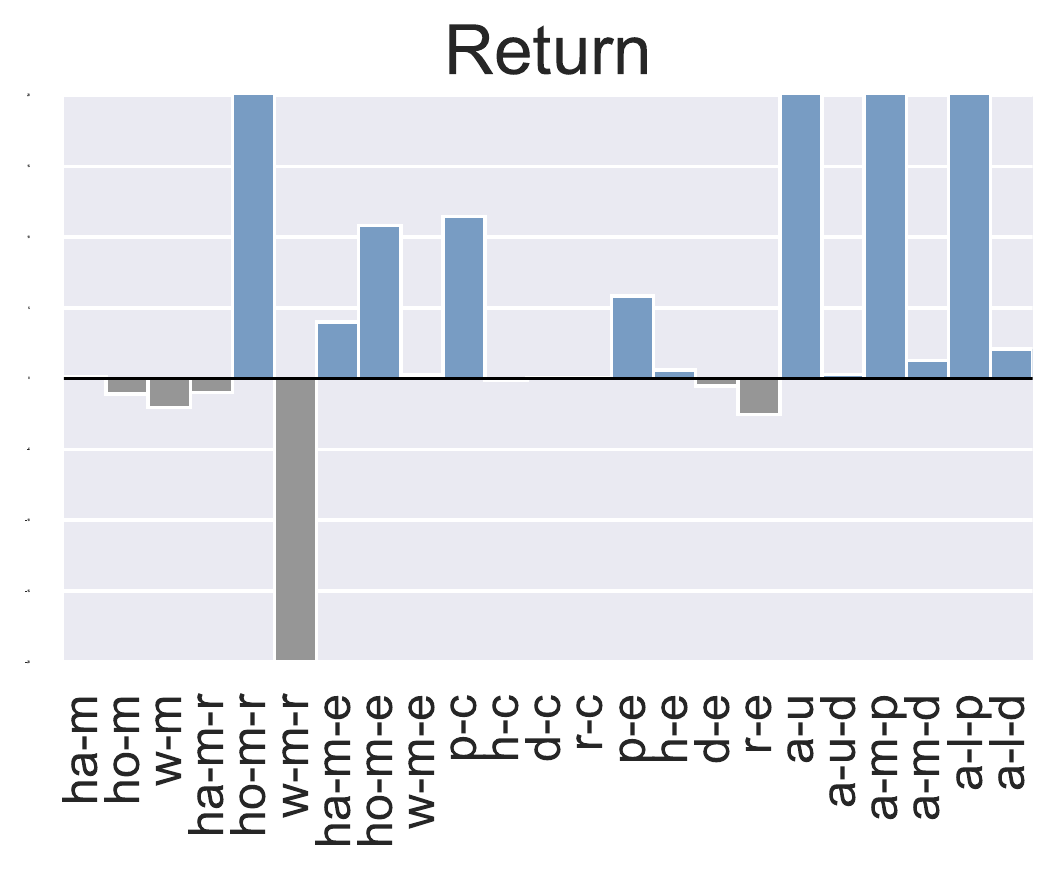}} \hspace{-6px}
  \subfigure{\includegraphics[height=73.1px]{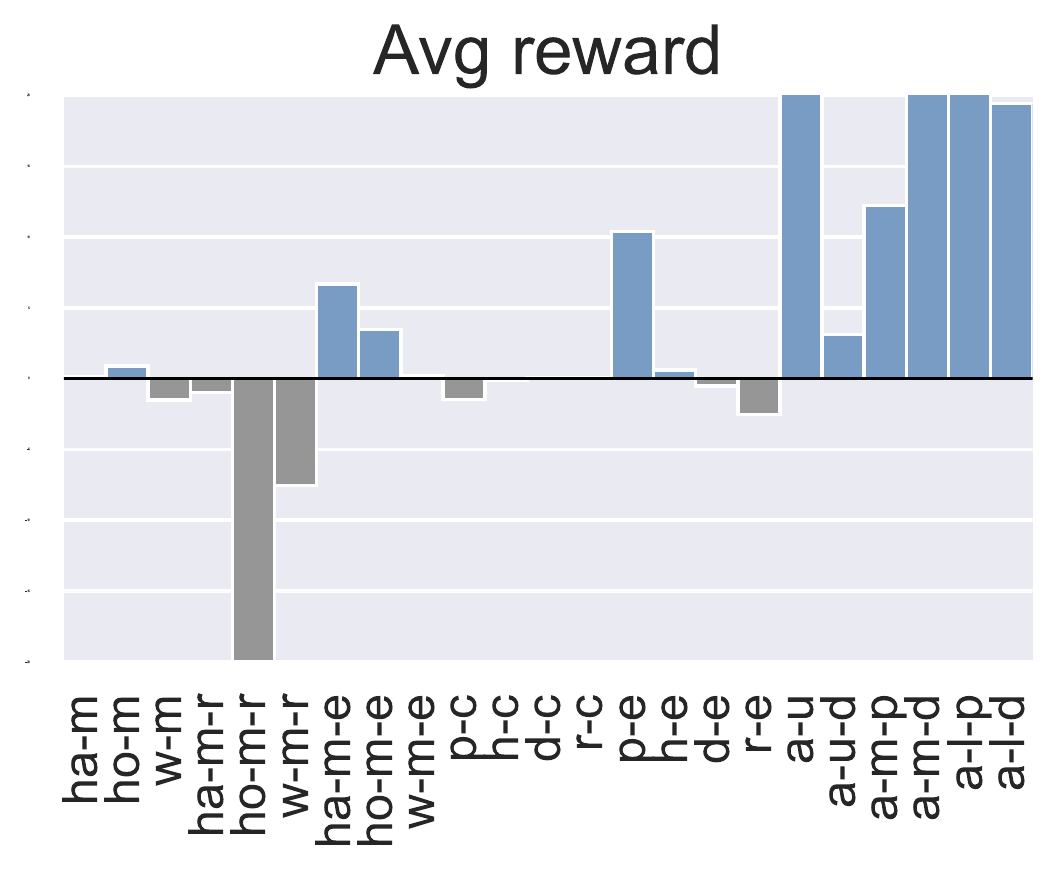}} \hspace{-6px}
  \subfigure{\includegraphics[height=73.1px]{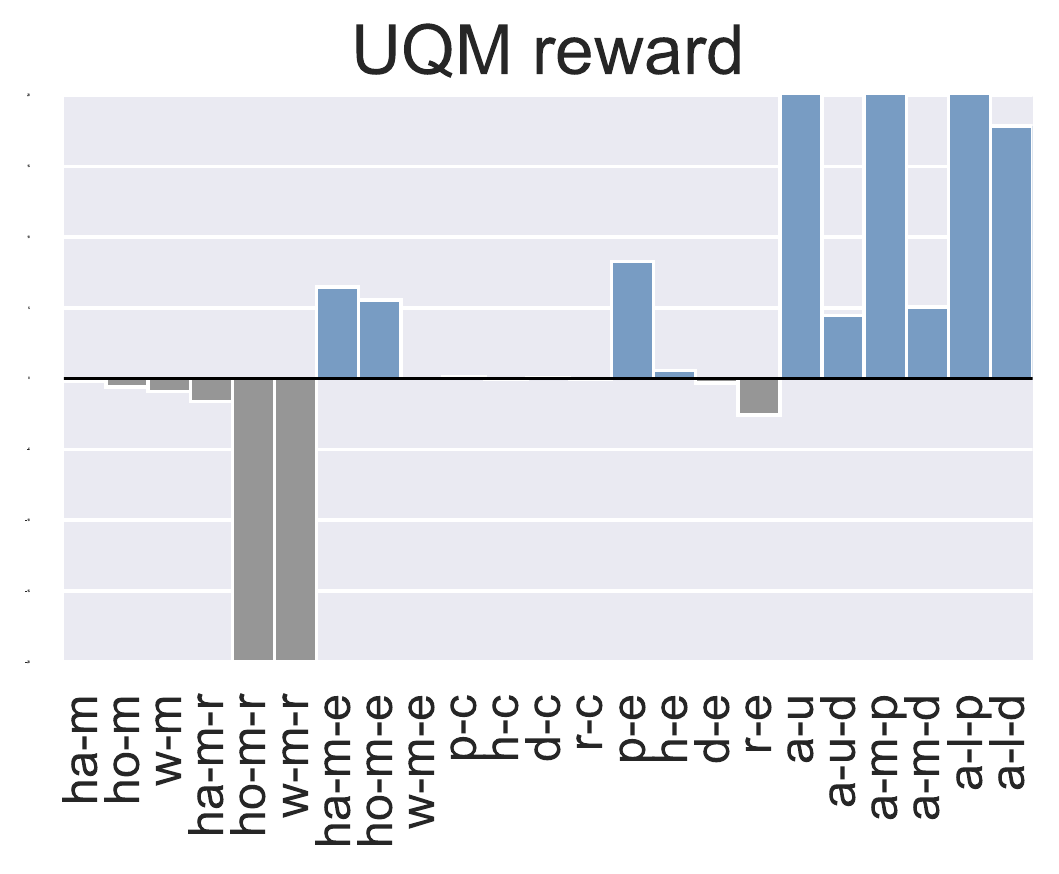}} \hspace{-6px}
  \subfigure{\includegraphics[height=73.1px]{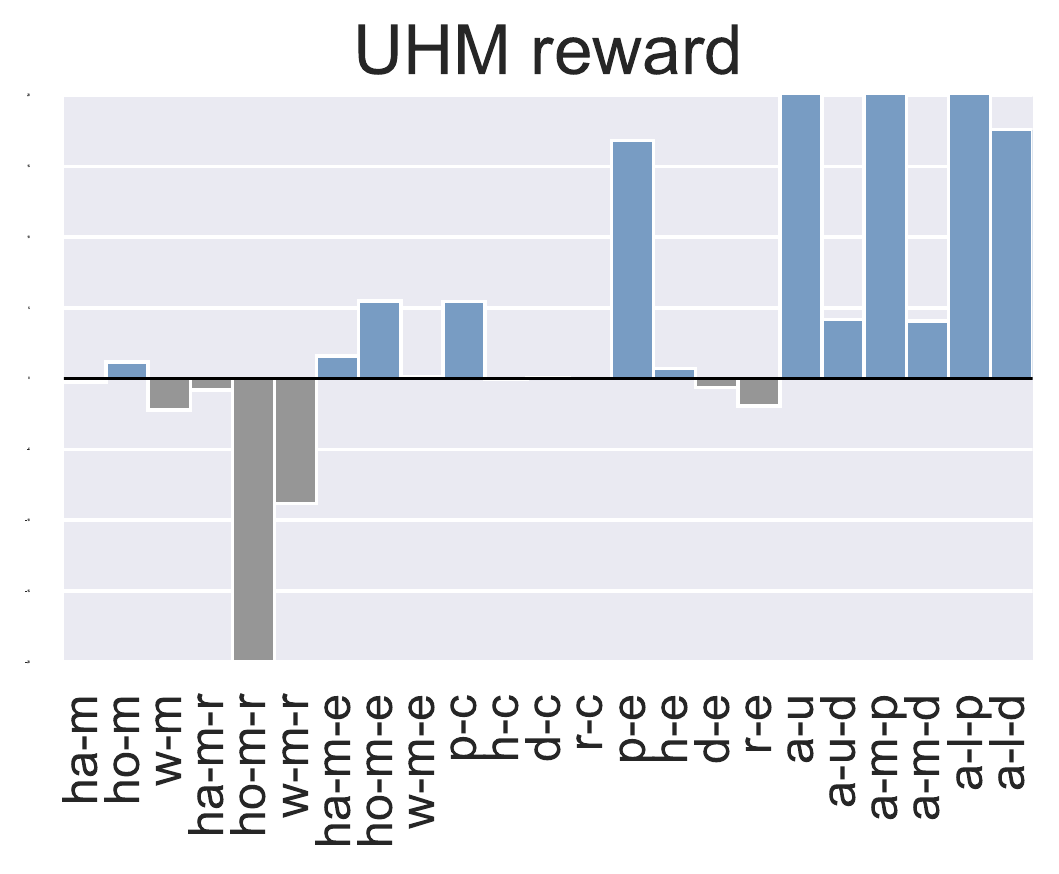}}
  \vspace{-6px}\\
  \subfigure{\includegraphics[height=79.5px]{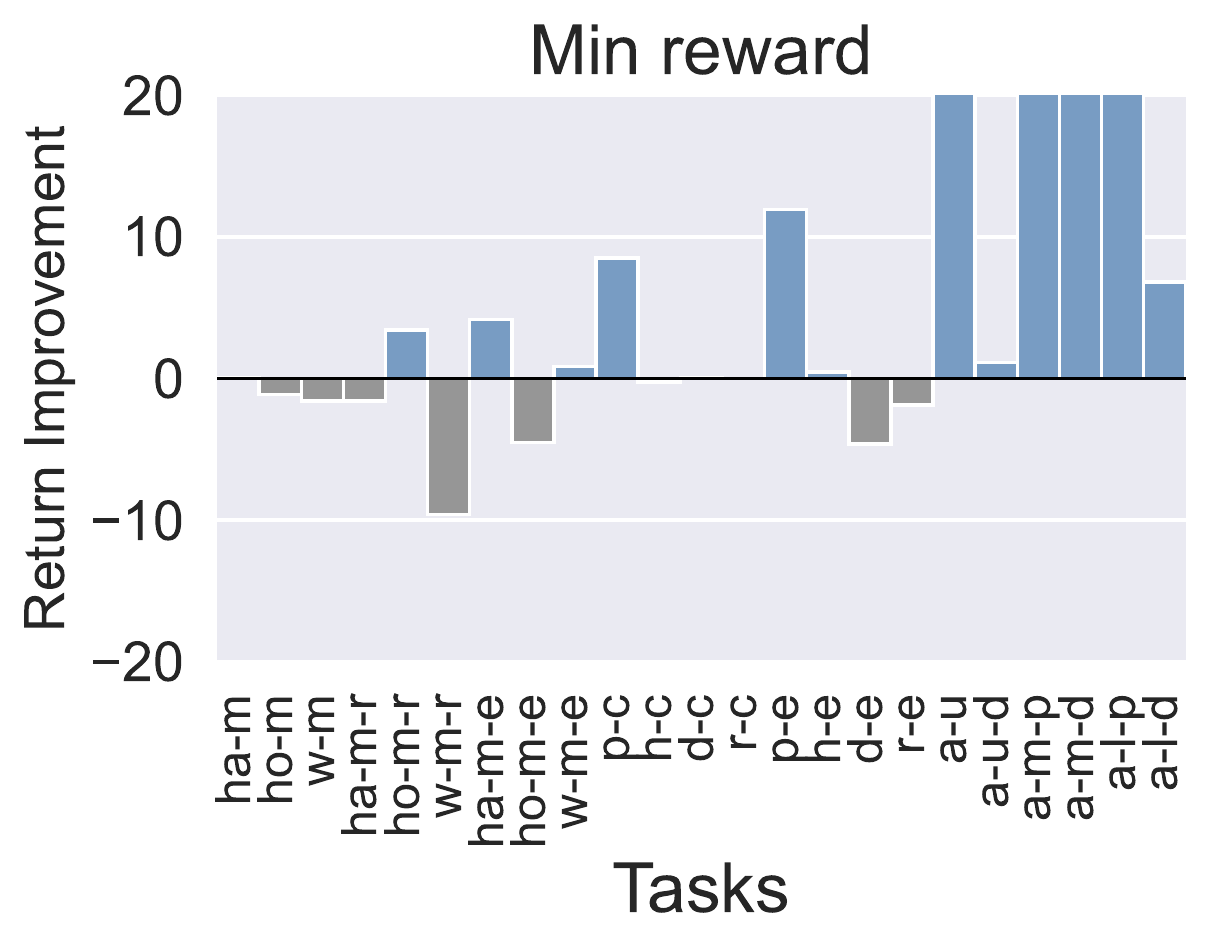}}\hspace{-3.4px}
  \subfigure{\includegraphics[height=79.5px]{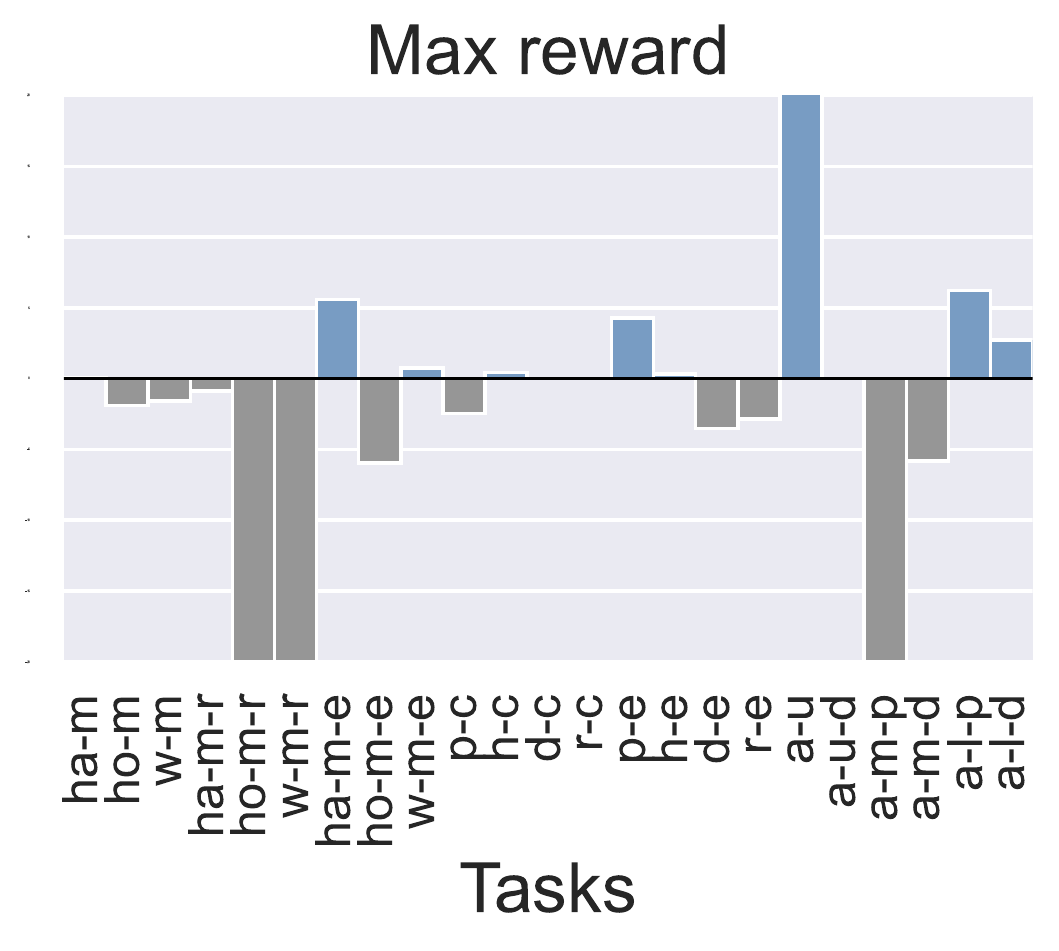}}\hspace{-3.4px}
  \subfigure{\includegraphics[height=79.5px]{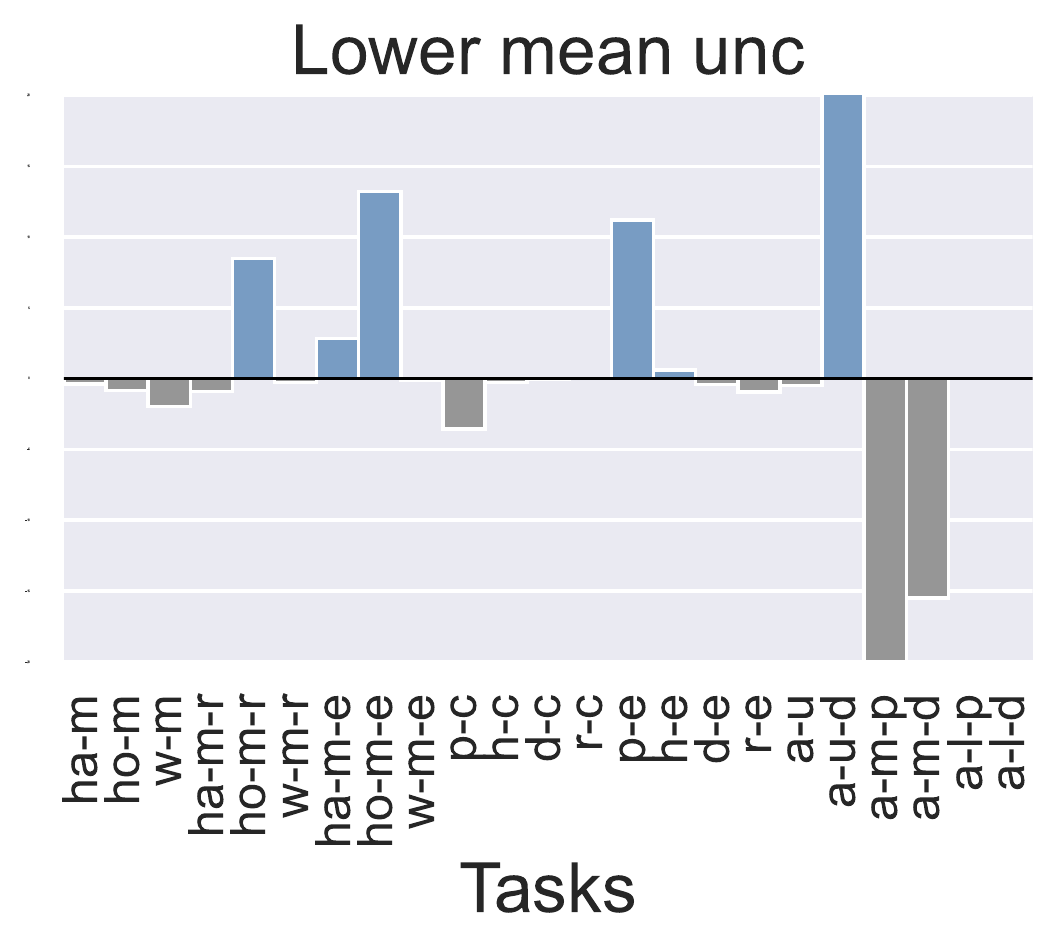}}\hspace{-3.4px}
  \subfigure{\includegraphics[height=79.5px]{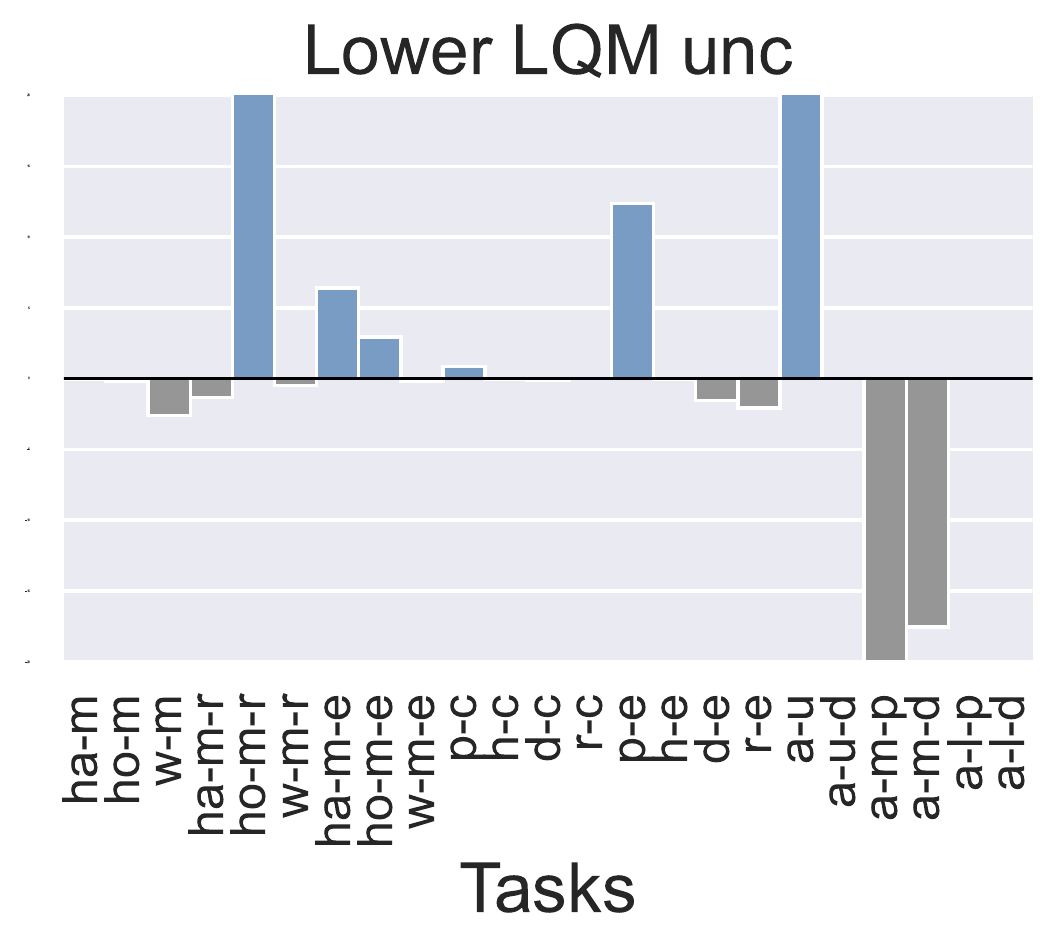}}\hspace{-3.4px}
  \subfigure{\includegraphics[height=79.5px]{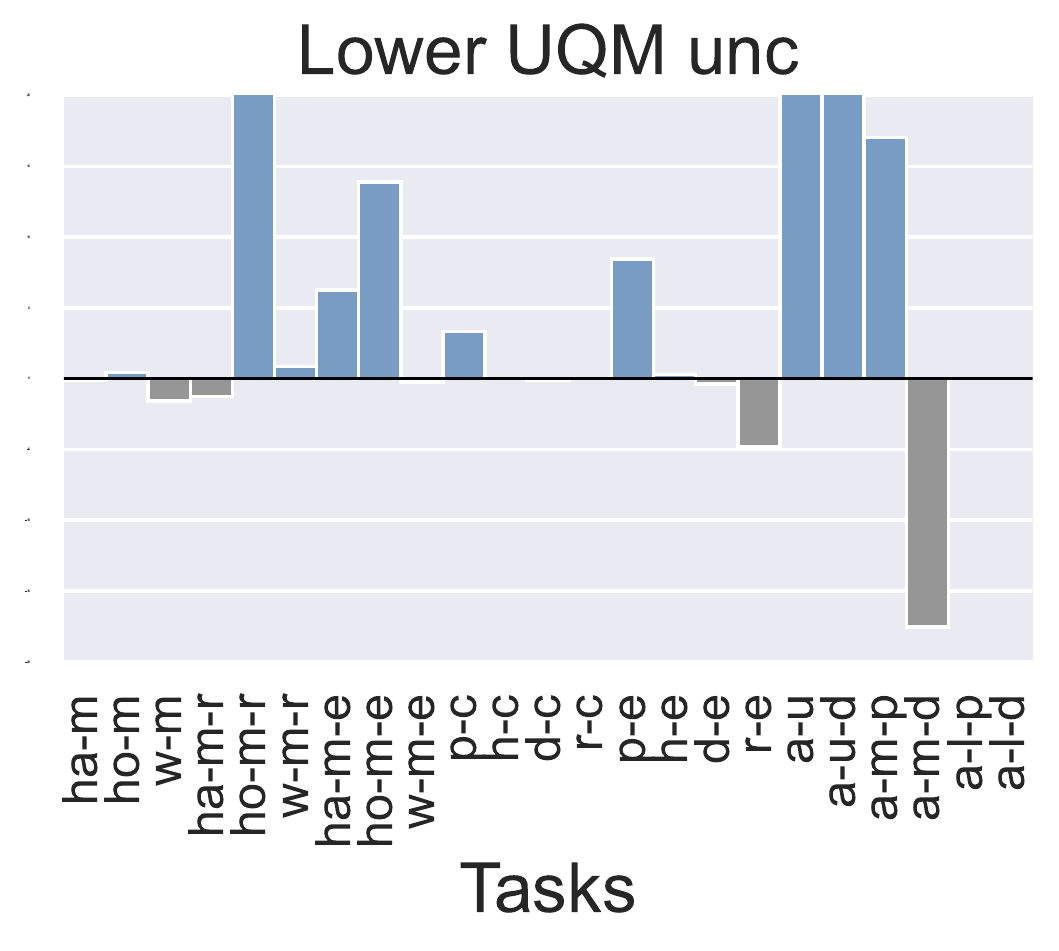}}
  \caption{Comparison of the performance of PTR compared to TD3+BC under 10 different trajectory priority metrics. In the figure we restrict the performance difference to a maximum of 20.}
  \label{fig:ptr}
  \vskip -0.2in
\end{figure*}

In this section, we evaluate PTR, which optimizes the trajectory sampling process of TR.
We implement two types of trajectory priority metrics based on trajectory quality and trajectory uncertainty. Figure~\ref{fig:ptr} shows the performance difference of various sampling techniques and baseline based on TD3+BC, and Table~\ref{tab:PTR} presents the sum of performance for datasets on each of the three environments. For detailed and overall results, please refer to the Appendix.

On sparse-reward datasets~(Antmaze and Adroit), it is advisable to prioritize higher quality trajectories during sampling. Those sampling approaches mitigate the instability of TD3+BC training and improves performance compared to uniform sampling of TR. Among them, prioritizing trajectories with higher mean rewards (\textit{Avg.} or \textit{UQM reward}) is more beneficial and stable. 
Furthermore, prioritizing trajectories without low rewards~(\textit{Min reward}) on the trajectory has a significant impact on some datasets and is more stable overall than other trajectory quality metrics on TD3+BC. 

On dense-reward datasets, prioritizing trajectories with lower uncertainty is reasonable to achieve stable and outstanding performance, as demonstrated by the last three subfigures in Figure~\ref{fig:ptr}. These metrics gain better or equivalent performance on Mujoco datasets. This highlights the ability of uncertainty to capture the characteristics of dense reward trajectories. The metric \textit{Lower UQM unc.}, which seeks trajectories with lower overall uncertainty, is the best performing metric in these cases.

Additionally, for dense reward tasks, when there are significant differences in rewards between trajectories, such as \verb|mujoco-medium-expert| dataset, prioritized sampling based on trajectory quality is also a good option, which can prioritize the use of higher-quality data and achieve better results. However, this approach may harm performance on the \verb|mujoco-medium-replay| dataset. On the other hand, estimating uncertainty of data with sparse reward signals can be challenging, and thus, the performance of uncertainty-based metrics is limited on sparse-reward datasets.

\begin{table}
    \centering
    \caption{The comparison of various priority metrics}
\vskip 0.15in
    \scriptsize
    \begin{tabular}{>{\raggedleft}m{3.925cm}>{\raggedright}m{0.8cm}>{\raggedright}m{0.8cm}>{\raggedright}m{1.0cm}}
        \toprule
        The dataset property & \textbf{TR}\\Uniform & \textbf{PTR} \\Quality & \textbf{PTR} \\Uncertainty\tabularnewline 
        \midrule
        Sparse reward trajectories & \ding{52} & \ding{52} & \ding{56}   \tabularnewline
        Dense reward trajectories & \ding{56} & - & \ding{52} \tabularnewline
        Many but not all high quality trajectories & - & \ding{52} & -   \tabularnewline
        \bottomrule
    \end{tabular}
    \label{tab:ptr_cmp}
\vskip -0.1in
\end{table}
In summary, as shown in Table~\ref{tab:ptr_cmp}, for tasks with sparse rewards, the basic trajectory sampling schema in TR can already show advantages, and it is also advisable to prioritize trajectories based on their quality, especially prioritizing based on \textit{UQM} or \textit{Avg. reward}, to gain better overall performance. Besides, for tasks with dense rewards, prioritizing trajectories of lower uncertainty is recommended, with \textit{Lower UQM unc.} being the best metric for greater benefits. If there are ample high-quality data in the dataset, prioritizing based on trajectory quality is also worth considering.

\subsection{Computational Cost Comparison}

We conduct experiments on a single machine with a NVIDIA GeForce 1080Ti 11GB GPU, and compare the computational cost of different sampling techniques of PTR based on TD3+BC in Table~\ref{tab:cost_cmp}. We record the average training time for each epoch (i.e., 1000 training steps).

The results indicate that trajectory-based sampling incurs a slightly higher computational cost, with an increase of 1 seconds per epoch compared to TD3+BC. This is primarily due to the maintenance of the available trajectories set during the sampling process. Besides, the uncertainty-based prioritized trajectory sampling requires less than 3 seconds more per epoch in contrast to quality-based sampling. This additional cost can be attributed to the continuous updating of the uncertainty value, which determines the sampling probability. These results imply that our PTR can extract more comprehensive information from limited data while maintaining a relatively low extra computational cost.

\begin{table}
    \centering
    \caption{Computational costs.}
\vskip 0.15in
    \scriptsize
    \begin{tabular}{>{\raggedright}m{2.5cm}>{\centering}m{2.5cm}}
        \toprule
        Sampling Method & Average Training Time \\(s/epoch)\tabularnewline 
        \midrule
        TD3+BC & 33.91  \tabularnewline
        TR & 35.24  \tabularnewline
        PTR~(quality) & 34.88  \tabularnewline
        PTR~(uncertainty) & 37.94  \tabularnewline
        \bottomrule
    \end{tabular}
    \label{tab:cost_cmp}
    \vskip -0.1in
\end{table}

%% file: sec/con.tex
\section{Conclusion and Limitation}

This paper investigates the effects of data sampling techniques, including trajectory-based backward sampling and priority-based trajectory sampling, on offline RL from the perspective of trajectories. We propose PTR to integrate these trajectory-based data sampling techniques as a plug-and-play replay memory module, which can be easily combined with any offline RL algorithm. Evaluation on D4RL datasets shows that trajectory-based backward sampling performs well, especially on sparse reward tasks. During the process of sampling trajectories, using reward mean, UQM, minimum value, or the reciprocal of uncertainty mean as metrics to guide probability sampling generally produces better results. This study aims to contribute and inspire further research on offline RL algorithms from the perspective of data sampling techniques. 

Nevertheless, this work has some limitations that need to be addressed in future research, such as the lack of theoretical discussion on the convergence of the proposed weighted target and the failure to improve performance on extremely sparse datasets like \verb|door-cloned-v1|, which may be due to insufficient refinement priority metrics. Moreover, guiding the process of collecting offline data based on our findings on trajectory quality and uncertainty is of great importance in real industrial scenarios.

%% file: sec/appendix.tex
\section{Experiment Details}
In this section, we provide implementation details of baseline algorithms and our proposed PTR.
\subsection{Implementation Details of baseline algorithms}
For a fair comparison and for a more convenient access to our proposed memory structure PTR, we have implemented and evaluated all the algorithms involved in this study on the basis of the CORL repository (\url{https://github.com/tinkoff-ai/CORL}) . 
\paragraph{TD3+BC}
    The critical hyper-parameter in TD3+BC is $\alpha$ value, which controls the weight of RL learning and behavior cloning process. The $\alpha$ is set 2.5 by default in the original paper in Mujoco tasks, and we find that the value of 2.5 works well only in Mujoco tasks. Thus, we conduct a hyperparameter search in the ranges (0.0001, 0.05, 0.25, 2.5, 25, 36, 50, 100) for Antmaze and Adroit datasets, and report the results in our results tables. Finally, the $\alpha$ values we find that bring the best performance are listed in Table~\ref{tab:hy-td3bc}. 
\begin{table}[h]
    \centering
    \caption{The $\alpha$ value for TD3+BC used in each datasets.}
    \vskip 0.15in
    \footnotesize
    \begin{tabular}{>{\raggedright}m{4.2cm}|>{\raggedleft}m{1.25cm}}
        \toprule
        \textbf{Datasets} & \textbf{$\alpha$ value}\tabularnewline 
        \midrule
        All Mujoco datasets \\antmaze-umaze-v0 & 2.5  \tabularnewline
        \midrule
        antmaze-umaze-diverse-v0\\antmaze-large-play-v0\\antmaze-large-diverse-v0 & 50  \tabularnewline
        \midrule
        antmaze-medium-play-v0 & 36  \tabularnewline
        \midrule
        antmaze-medium-diverse-v0 & 25  \tabularnewline
        \midrule
        pen-cloned-v1 & 0.0001  \tabularnewline
        \midrule
        hammer-cloned-v1 \\ relocate-expert-v1 & 0.01  \tabularnewline
        \midrule
        door-cloned-v1 \\ relovate-cloned-v1 \\ pen-expert-v1\\ hammer-expert-v1\\ door-expert-v1 & 0.25  \tabularnewline
        \bottomrule
    \end{tabular}
    \label{tab:hy-td3bc}
    \vskip -0.1in
\end{table}

\paragraph{IQL}
We use the hyper-parameters that alines with their original paper, and our reported results on Mujoco and Antmaze are close to the reported results. Specifically, we use $\tau = 0.9$ and $\beta = 10.0$ for Antmaze tasks, $\tau=0.7$ and $\beta = 3.0$ for Mujoco tasks, and $\tau=0.7$ and $\beta=0.5$ for Adroit datasets.

\paragraph{EDAC}
We set the weight of the ensemble gradient diversity term $\eta$ following the original paper. Note that $N$ is original set as 10 for \verb|halfcheetah| and \verb|walker2d| datasets and 50 for \verb|hopper| dataset. Rather, considering the computational cost, we set it as 3 for all \verb|mujoco-medium-replay-v2| datasets, and 10 for others. Our reproduced results in most of Mujoco datasets are similar with their reported results. In Adroit datasets, despite obtaining poor results due to inappropriate $N$ values, we are still able to showcase the subtle yet clearly discernible advantage of TR. Note that since the authors have not given the appropriate hyperparameter settings for EDAC on Antmaze in the original paper, and our practice of fine tuning the hyperparameters did not yield good result, we omit the results of EDAC on Antmaze.

\subsection{Implementation Details of our proposed methods}
We also implement TR and PTR based on CORL as a plug-in that can be flexibly combined with any baseline algorithm to replace the traditional state-transition-based replay memory with our proposed trajectory-based memory - PTR.

\begin{table}[t]
    \centering
    \caption{The $\beta$ value for weighted target used in each datasets.}
    \vskip 0.15in
    \footnotesize
    \begin{tabular}{>{\raggedright}m{4.2cm}|>{\raggedleft}m{1.25cm}}
        \toprule
        \textbf{Datasets} & \textbf{$\beta$ value}\tabularnewline 
        \midrule
        halfcheetah-medium-v2\\halfcheetah-medium-replay-v2\\halfcheetah-medium-expert-v2\\hopper-medium-replay-v2\\hopper-medium-expert-v2  & 1.0  \tabularnewline \midrule
        hopper-medium-v2 \\ walker2d-medium-v2 & 0.0 \tabularnewline  \midrule
        walker2d-medium-replay-v2 \\ walker2d-medium-expert-v2 & 0.5 \tabularnewline \midrule 
        antmaze-umaze-v0 & 0.95 \tabularnewline 
antmaze-umaze-diverse-v0 & 0.25\tabularnewline 
antmaze-medium-play-v0 & 0.98 \tabularnewline 
antmaze-medium-diverse-v0 & 0.75 \tabularnewline 
antmaze-large-play-v0 & 0.75 \tabularnewline 
antmaze-large-diverse-v0 & 0.98 \tabularnewline 
\midrule
pen-cloned-v1   & 0.5\tabularnewline
hammer-cloned-v1 & 1.0\tabularnewline
door-cloned-v1 & 1.0\tabularnewline
relocate-cloned-v1 & 0.05\tabularnewline
pen-expert-v1 & 0.5\tabularnewline
hammer-expert-v1 &0.5 \tabularnewline
door-expert-v1 & 0.0\tabularnewline
relocate-expert-v1 &0.0 \tabularnewline
        \bottomrule
    \end{tabular}
    \label{tab:hy-beta}
    \vskip -0.1in
\end{table}

\begin{figure*}[t]
\vskip 0.2in
  \centering
  \subfigure[Dense reward trajectories]{\includegraphics[height=95px]{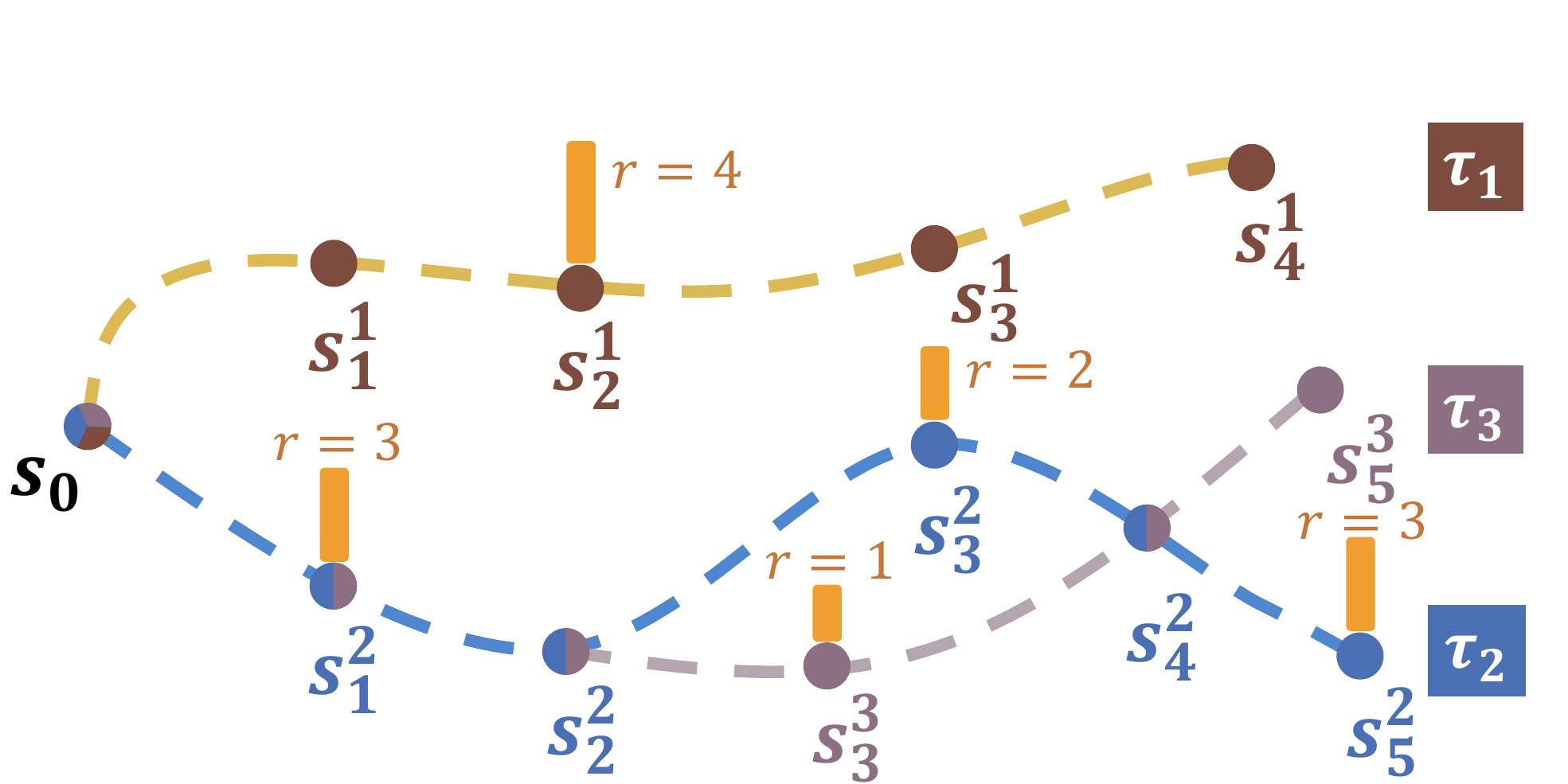}} 
  \hspace{5px}
  \subfigure[Sparse reward trajectories]{\includegraphics[height=95px]{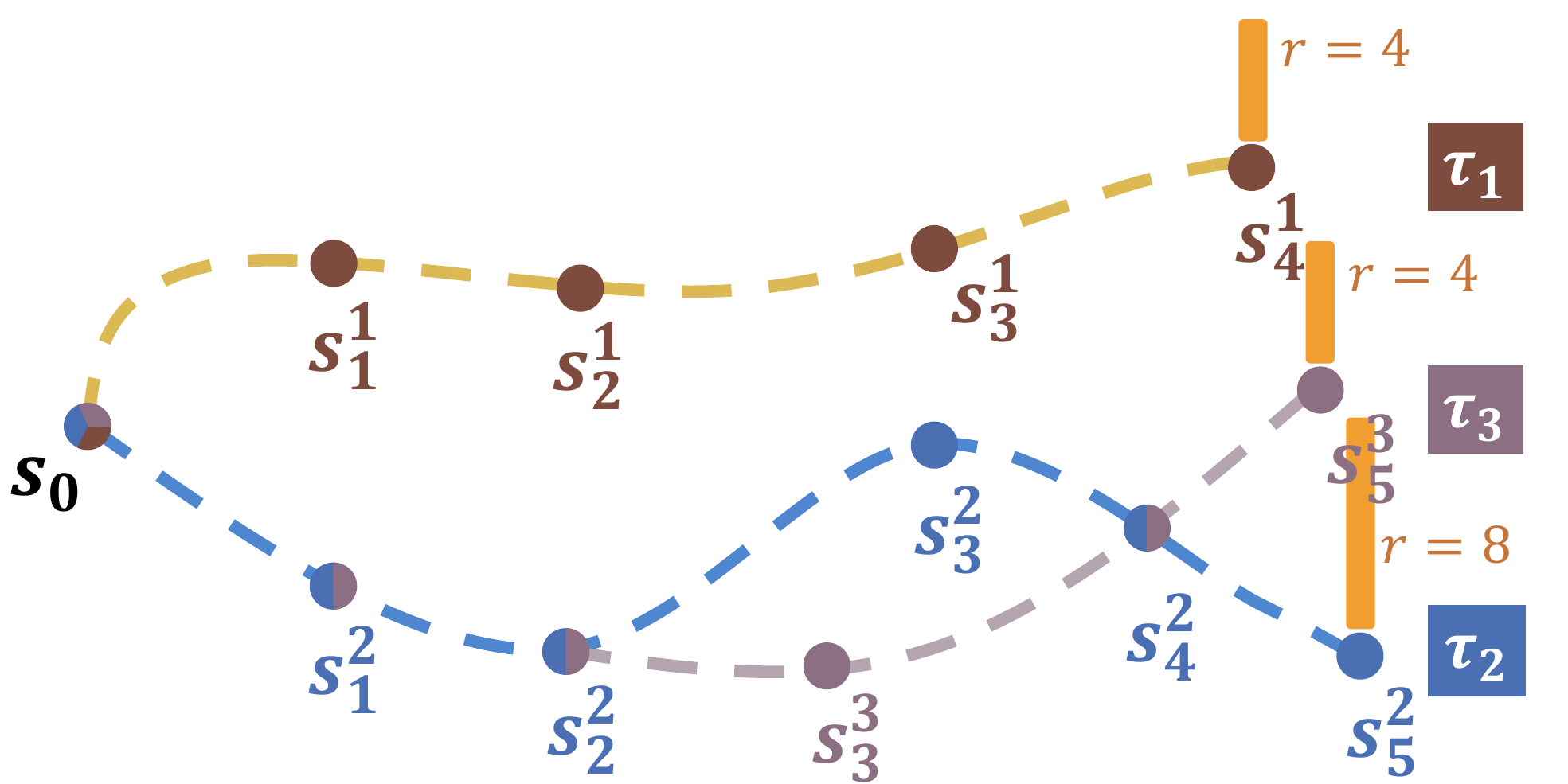}} 
  \caption{The motivating example on finite data. }
  \label{fig:fig1_app}
  \vskip -0.2in
\end{figure*}
\paragraph{TR}

The basic TR is implemented for storing and sampling the state transitions in trajectories' order. In TR, we first need to store the offline dataset in the form of trajectories so that we can access the specific state information on a particular trajectory by indexing the trajectory number and indexing the time step on the trajectory. For the available trajectories set and the sampled trajectories set in Figure~\ref{fig:TR-based-backward}, we maintain two arrays to store the corresponding trajectory numbers for a more convenient and less expensive implementation. We encapsulate the sampling process into the plug-in \verb|TrajectoryReplay| so that we can access different offline training algorithms with minimal cost. Specifically, all we need to do is to define TR and load offline data in TR by the following code, and call its sampling function when we need the batch data for training.
\begin{verbatim}
memory = TrajectoryReplay()
memory.load_offline_dataset(dataset)
    ...
batch_data = memory.sample(batch_size)
\end{verbatim}

We introduce an additional critic target calculation in Sec.~\ref{sec:weighted_target}. In this, we maintain a $Q$ table on the current trajectory to store the computed values of $Q$\_target at time-step $t+1$. During this process, we introduce the hyperparameter - the only hyperparameter involved in this research - $\beta$. In the specific experiments, we find that a suitable value of $\beta$ can bring a very significant performance improvement, and the value of $\beta$ on different datasets corresponding to our final reported results is shown in Table~\ref{tab:hy-beta}.

\begin{table*}[t]
  \caption{Normalized average returns of TR and baselines on Gym Mujoco, Antmaze, Adroit tasks in D4RL. The averaged performance and standard deviation of 3 runs are reported.}
  \vskip 0.15in
  \label{tab:tab1_full}
  \renewcommand{\arraystretch}{0.96}
  \centering
  \scriptsize
\begin{tabular}{>{\raggedright}m{1.6cm}|>{\raggedleft}m{1.5cm}>{\raggedleft}m{1.5cm}|>{\raggedleft}m{1.5cm}>{\raggedleft}m{1.5cm}|>{\raggedleft}m{1.5cm}>{\raggedleft}m{1.5cm}}
\toprule
\textbf{Task Name} & \textbf{TD3+BC} & \textbf{TD3+BC(TR)} & \textbf{EDAC} & \textbf{EDAC(TR)} & \textbf{IQL} & \textbf{IQL(TR)}\tabularnewline
\midrule
halfcheeh-m & 48.20$\pm$0.37 & 48.13$\pm$0.38 & 66.73$\pm$2.07 & 64.01$\pm$2.42 & 48.24$\pm$0.13 & 47.96$\pm$0.19\tabularnewline
hopper-m & 60.47$\pm$5.76 & 59.96$\pm$2.01 & 78.19$\pm$18.81 & 89.92$\pm$14.41 & 58.45$\pm$2.58 & 61.81$\pm$4.82\tabularnewline
walker2d-m & 84.12$\pm$1.68 & 82.95$\pm$2.51 & 71.60$\pm$31.38 & 87.16$\pm$2.28 & 80.19$\pm$6.44 & 78.79$\pm$2.88\tabularnewline
halfcheetah-m-r & 44.76$\pm$0.31 & 43.91$\pm$0.79 & 64.31$\pm$1.39 & 57.79$\pm$4.26 & 43.48$\pm$0.58 & 43.22$\pm$0.57\tabularnewline
hopper-m-r & 50.60$\pm$23.04 & 49.83$\pm$18.01 & 97.16$\pm$9.45 & 32.01$\pm$5.47 & 84.21$\pm$17.45 & 80.86$\pm$24.64\tabularnewline
walker2d-m-r & 77.83$\pm$16.17 & 36.14$\pm$33.99 & 25.55$\pm$24.39 & 5.39$\pm$5.89 & 75.93$\pm$8.97 & 79.35$\pm$8.27\tabularnewline
halfcheetah-m-e & 89.26$\pm$4.48 & 93.63$\pm$1.04 & 71.72$\pm$8.70 & 68.87$\pm$15.288 & 92.06$\pm$4.398 & 94.82$\pm$0.91\tabularnewline
hopper-m-e & 94.01$\pm$10.39 & 105.50$\pm$6.16 & 28.44$\pm$7.211 & 28.65$\pm$10.82 & 96.73$\pm$11.81 & 72.74$\pm$28.02\tabularnewline
walker2d-m-e & 110.57$\pm$0.68 & 110.51$\pm$0.37 & 114.00$\pm$1.02 & 112.64$\pm$1.05 & 112.12$\pm$0.68 & 110.44$\pm$0.34\tabularnewline
\midrule
\textbf{Total} & \textbf{659.81} & 630.54 & \textbf{617.71} & 546.44 & \textbf{691.41} & 669.99 \tabularnewline
\midrule
\midrule
antmaze-u & 32.00$\pm$52.01 & 91.53$\pm$3.36 &  - & -& 66.70$\pm$13.76 & 76.23$\pm$4.67\tabularnewline
antmaze-u-d & 0.00$\pm$0.00 & 0.00$\pm$0.00 & - & -& 50.33$\pm$7.94 & 48.10$\pm$11.08\tabularnewline
antmaze-m-p & 35.65$\pm$37.50 & 44.20$\pm$38.70 & - & - & 66.80$\pm$9.87 & 65.93$\pm$10.36\tabularnewline
antmaze-m-d & 17.53$\pm$27.32 & 40.40$\pm$37.49 &  - & -& 71.30$\pm$5.35 & 64.93$\pm$21.52\tabularnewline
antmaze-l-p & 0.00$\pm$0.00 & 32.37$\pm$14.08 & - & - & 37.40$\pm$13.50 & 46.53$\pm$3.29\tabularnewline
antmaze-l-d & 0.00$\pm$0.00 & 15.17$\pm$7.72 &  - & - & 37.37$\pm$21.83 & 54.73$\pm$8.49\tabularnewline
\midrule
\textbf{Total} & 98.36 & \textbf{223.67} &   - & -  & 329.90 & \textbf{356.47} \tabularnewline
\midrule
\midrule
pen-cloned & 64.21$\pm$31.82 & 71.83$\pm$32.42 & 0.12$\pm$6.12 & 1.02$\pm$4.24 & 69.22$\pm$5.38 & 69.40$\pm$8.06\tabularnewline
hammer-cloned & 0.62$\pm$0.51 & 0.93$\pm$1.05 & 0.21$\pm$0.01 & 0.22$\pm$0.37 & 2.58$\pm$1.90 & 2.03$\pm$1.85\tabularnewline
door-cloned & -0.11$\pm$1.14 & -0.08$\pm$0.05 & -0.35$\pm$0.13 & -0.33$\pm$0.02 & -0.04$\pm$0.04 & 0.03$\pm$0.14\tabularnewline
relocate-cloned & -0.26$\pm$0.04 & -0.24$\pm$0.03 & -0.05$\pm$0.08 & -0.17$\pm$0.10 & -0.05$\pm$0.02 & -0.06$\pm$0.13\tabularnewline
pen-expert & 136.77$\pm$18.28 & 138.74$\pm$5.53 & -1.07$\pm$1.52 & 6.36$\pm$8.84 & 136.14$\pm$6.00 & 136.60$\pm$6.48\tabularnewline
hammer-expert & 128.58$\pm$0.45 & 128.60$\pm$0.38 & 0.25$\pm$0.15 & 0.39$\pm$0.54 & 127.43$\pm$0.41 & 127.36$\pm$0.42\tabularnewline
door-expert & 105.75$\pm$0.78 & 105.07$\pm$1.26 & 3.63$\pm$6.33 & 16.27$\pm$28.25 & 105.68$\pm$0.93 & 105.61$\pm$0.19\tabularnewline
relocate-expert & 104.84$\pm$3.54 & 101.11$\pm$3.19 & -0.35$\pm$0.00 & -0.35$\pm$0.00 & 104.73$\pm$0.83 & 102.56$\pm$1.01\tabularnewline
\midrule
\textbf{Total} & \textbf{540.41} & \textbf{545.95} &  2.38 & 23.41 & \textbf{545.70} & \textbf{543.52}\tabularnewline
\bottomrule
\end{tabular}
\vskip -0.1in
\end{table*}

\paragraph{PTR}
The two types of trajectory priority metrics are based on trajectory quality and trajectory uncertainty, respectively. The difference between them lies in the fact that the former is an inherent characteristic of the trajectory itself and does not change with training, while the latter is dynamic. Therefore, for the former, we only need to maintain a metric array related to trajectory priority while maintaining trajectory information, and then perform probability sampling when sampling trajectories. For trajectory sampling based on uncertainty, we additionally need to update the uncertainty metrics of the trajectory after each complete trajectory sampling. The updating process should be performed to ensure that the priority ranking of trajectories based on uncertainty is dynamically adjusted and reflects the latest situation.

\subsection{Experimental Settings}
For all experiments, we train 1000 epochs (1000 training steps for each epoch), i.e., $10^6$ training steps in total. 
We report the normalized d4rl score: $\text{normalized score} = 100 * \frac{\text{score} - \text{random score}}{\text{expert score} - \text{random score}}$, following the function interface \verb|get_normalized_score()| provided by D4RL\footnote{\url{https://github.com/Farama-Foundation/D4RL/blob/master/d4rl/offline_env.py}}. Besides, for the fair evaluation and avoid the effect of instability, we are not reporting the evaluation results of the last epoch, rather the averaged evaluation results of the last 5 epochs, i.e., the last 50 evaluation episodes.

\section{The Details for the Illustrative Example on Finite Data}

Figure~\ref{fig:fig1} illustrates a simple example on finite data with only 3 trajectories. These trajectories all start at state $s_0$, and the objective this example is to estimate the maximum $Q$-value at state $s_0$. Figure~\ref{fig:fig1_app} shows the two cases we mentioned before. In dense reward case, the reward can be obtained in several states. Nevertheless, in sparse reward case, the reward can only be obtained in the end of each trajectory.

The reported curves in Figure~\ref{fig:fig1} represent the learned maximum $Q$-value at state $s_0$. Note that the undiscounted cumulative rewards of these 3 trajectories are 4, 8, 4, respectively. Hence, considering the discount factor~(0.99), the estimated maximum $Q$-value at state $s_0$ should converge to a value a little less than 8, exactly 7.76 for dense case and 7.61 for sparse case.

\section{Additional Experimental Results}
In this section, we show the additional and detailed experimental results.
\subsection{Detailed Results of TR}
Table~\ref{tab:tab1} shows the total performance of baseline algorithms and TR, based on TD3+BC, EDAC, and IQL. For more details, in Table~\ref{tab:tab1_full}, we show the full results of each dataset. As can be seen from the results, the trajectory-based sampling process of TR provides advantages on almost all Antmaze datasets. Performance gains are also evident on \verb|pen-cloned-v1| and \verb|pen-expert-v1|. Yet on the other Adroit datasets, TR maintains the performance level of the baseline algorithm - seemingly reaching a performance bottleneck. On the dense reward Mujoco, the performance improvement of TR based on TD3+BC is more noticeable on the \verb|-medium-expert-v2| dataset, and the performance improvement of TR based on EDAC is more noticeable on the \verb|-medium-v2| dataset, however, on the \verb|-medium-replay-v2| dataset, both TD3+BC-based and EDAC-based TR differently impair the performance to a certain extent, which demonstrates the instability of the most basic TR on the dense reward dataset.

\subsection{Detailed Results of Prioritized Trajectory Replay}
In Table~\ref{tab:apptd3} and Table~\ref{tab:appiql}, we show the original results of PTR on D4RL tasks, which are the extended tables of Table~\ref{tab:PTR}. The details of the results make some conclusions more evident. For dense reward tasks, uncertainty-based prioritized trajectory sampling shows more stable performance, especially with L\_UQM\_u (i.e., sampling trajectories with higher priority for the trajectories whose top 25\% uncertainty values are smaller). This is consistent with researchers' consistent understanding that data with lower uncertainty is more useful for offline reinforcement learning training. For sparse reward tasks, priority trajectory sampling based on trajectory quality performs better, especially with UQM\_r and Avg\_r, indicating that higher quality trajectories are more helpful for learning. In addition, we verified that the mean reward is a more reasonable metric than the return, and we hope this will be valued in future research.

\subsection{The results of prioritizing higher uncertainty trajectories}
We also verify the effect of prioritizing trajectories with higher uncertainty values. Not surprisingly, this approach significantly damages performance. The extrapolation error caused by higher uncertainty is severe. However, in online RL, higher uncertainty is often regarded as a guide for exploration and can be used to help the agent obtain more useful data. Therefore, we propose three metrics, \textit{Higher mean unc.}, \textit{Higher LQM unc.}, and \textit{Higher UQM unc.}, which we hope will be studied in online RL in future work.

\begin{sidewaystable}
  \centering
  \vspace{3.6in}
  \caption{Normalized average returns of TD3+BC.}
  \vspace{0.15px}
  \label{tab:apptd3}
  \scriptsize
\begin{tabular}{>{\raggedright}m{1.6cm}|>{\raggedleft}m{1.0cm}>{\raggedleft}m{1.0cm}>{\raggedleft}m{1.0cm}>{\raggedleft}m{1.0cm}>{\raggedleft}m{1.0cm}>{\raggedleft}m{1.0cm}>{\raggedleft}m{1.0cm}>{\raggedleft}m{1.0cm}>{\raggedleft}m{1.0cm}>{\raggedleft}m{1.0cm}>{\raggedleft}m{1.0cm}>{\raggedleft}m{1.0cm}>{\raggedleft}m{1.0cm}>{\raggedleft}m{1.0cm}}
\toprule
Task Name & TD3+BC & TD3+BC\\(TR) & TD3+BC\\(Return) & TD3+BC\\(Avg\_r) & TD3+BC\\(UQM\_r) & TD3+BC\\(UHM\_r) & TD3+BC\\(Min\_r) & TD3+BC\\(Max\_r) & TD3+BC\\(L\_mean) & TD3+BC\\(L\_LQM) & TD3+BC\\(L\_UQM) & TD3+BC\\(H\_mean) & TD3+BC\\(H\_LQM) & TD3+BC\\(H\_UQM)\tabularnewline
\midrule
halfcheeh-m & 48.2$\pm$0.4 & 48.1$\pm$0.4 & 48.4$\pm$0.2 & 48.4$\pm$0.2 & 48.0$\pm$0.3 & 48.0$\pm$0.2 & 48.3$\pm$0.5 & 48.3$\pm$0.2 & 47.8$\pm$0.5 & 48.1$\pm$0.4 & 48.1$\pm$0.3 & 47.8$\pm$0.2 & 47.7$\pm$0.3 & 47.7$\pm$0.4\tabularnewline
hopper-m & 60.5$\pm$5.8 & 60.0$\pm$2.1 & 59.4$\pm$2.6 & 61.4$\pm$5.9 & 59.9$\pm$3.8 & 61.6$\pm$4.2 & 59.3$\pm$6.3 & 58.6$\pm$2.8 & 59.6$\pm$2.5 & 60.3$\pm$4.3 & 60.9$\pm$4.4 & 54.8$\pm$5.0 & 56.0$\pm$6.1 & 56.6$\pm$5.3\tabularnewline
walker2d-m & 84.1$\pm$1.7 & 82.9$\pm$2.5 & 82.1$\pm$1.6 & 82.6$\pm$2.0 & 83.2$\pm$2.2 & 81.9$\pm$2.7 & 82.5$\pm$2.5 & 82.5$\pm$2.6 & 82.1$\pm$2.0 & 81.5$\pm$2.6 & 82.5$\pm$2.5 & 78.9$\pm$3.9 & 82.0$\pm$2.9 & 78.4$\pm$7.1\tabularnewline
halfcheetah-m-r & 44.8$\pm$0.3 & 43.9$\pm$0.8 & 43.8$\pm$0.5 & 43.8$\pm$0.5 & 43.1$\pm$0.7 & 44.0$\pm$0.8 & 43.2$\pm$0.7 & 43.9$\pm$0.3 & 43.8$\pm$0.4 & 43.4$\pm$0.6 & 43.5$\pm$0.5 & 42.9$\pm$0.6 & 43.1$\pm$0.7 & 43.1$\pm$0.7\tabularnewline
hopper-m-r & 50.6$\pm$23.0 & 49.8$\pm$18.1 & 81.8$\pm$14.6 & 29.1$\pm$8.8 & 26.5$\pm$4.1 & 25.0$\pm$5.2 & 54.0$\pm$24.9 & 24.4$\pm$1.4 & 59.1$\pm$18.4 & 76.4$\pm$20   & 71.2$\pm$13.7 & 29.9$\pm$13.8 & 35.9$\pm$18.3 & 27.7$\pm$6.9\tabularnewline
walker2d-m-r & 77.8$\pm$16.1 & 36.1$\pm$34.0 & 7.9$\pm$7.2 & 70.3$\pm$16.2 & 55.1$\pm$46.4 & 69.0$\pm$16.2 & 68.2$\pm$9.0 & 53.0$\pm$39.5 & 77.5$\pm$11.9 & 77.4$\pm$8.5  & 78.7$\pm$4.9  & 12.2$\pm$14.2 & 12.7$\pm$7.1  & 6.4$\pm$9.3\tabularnewline
halfcheetah-m-e & 89.3$\pm$4.5 & 93.6$\pm$1.0 & 93.2$\pm$2.1 & 95.9$\pm$0.5 & 95.7$\pm$1.3 & 90.9$\pm$7.4 & 93.4$\pm$2.0 & 94.9$\pm$3.1 & 92.1$\pm$5.8  & 95.7$\pm$1.5  & 95.5$\pm$1.6  & 80$\pm$5.7    & 80.3$\pm$5.8  & 79.4$\pm$6\tabularnewline
hopper-m-e & 94.0$\pm$10.4 & 105.5$\pm$6.2 & 104.8$\pm$8.8 & 97.5$\pm$10.7 & 99.5$\pm$10.5 & 99.5$\pm$11.1 & 89.5$\pm$16.6 & 88.1$\pm$9.9 & 107.2$\pm$4.4 & 97$\pm$19.5   & 107.9$\pm$4.5 & 85.2$\pm$16.2 & 86.5$\pm$15.5 & 86.7$\pm$14.4 \tabularnewline
walker2d-m & 110.6$\pm$0.7 & 110.5$\pm$0.4 & 110.8$\pm$0.4 & 110.8$\pm$0.5 & 110.6$\pm$0.9 & 110.7$\pm$0.5 & 111.4$\pm$0.8 & 111.3$\pm$0.4 & 110.5$\pm$0.3 & 110.4$\pm$0.4 & 110.3$\pm$0.3 & 110.1$\pm$1.6 & 108.2$\pm$4.2 & 105.3$\pm$6.7\tabularnewline
\midrule
Total & 659.81 & 630.54 & 632.17 & 639.62 & 621.69 & 630.57 & 649.92 & 604.87 &\textbf{ 679.88} & \textbf{690.18} & \textbf{698.59} & 541.69 & 552.16 & 531.33 \tabularnewline
\midrule
\midrule
antmaze-u          & 32.0$\pm$52.0 & 91.5$\pm$3.4  & 93.3$\pm$1.9  & 93.6$\pm$1.9  & 93.6$\pm$2.9  & 90.0$\pm$3.8  & 94.4$\pm$2.8  & 64.9$\pm$47.8 & 31.5$\pm$24.8 & 72.5$\pm$30.6 & 81.8$\pm$6.1  & 48.2$\pm$9.7 & 93.6$\pm$3.4  & 27.4$\pm$28.8 \tabularnewline
antmaze-u-d  & 0.0$\pm$0.0   & 0.0$\pm$0.0   & 0.3$\pm$0.5   & 3.1$\pm$5.3   & 4.5$\pm$7.7   & 4.2$\pm$7.0   & 1.1$\pm$2.0   & 0.0$\pm$0.0   & 22.9$\pm$39.6 & 0.0$\pm$0.0   & 37.1$\pm$34.0 & 2.8$\pm$4.9  & 4.1$\pm$4.9   & 2.1$\pm$3.7   \tabularnewline
antmaze-m-p    & 35.7$\pm$37.5 & 44.2$\pm$38.7 & 75.8$\pm$4.5  & 47.9$\pm$41.5 & 70.4$\pm$8.5  & 66.7$\pm$11.6 & 75.6$\pm$12.8 & 10.1$\pm$17.6 & 0.0$\pm$0.0   & 0.0$\pm$0.0   & 52.7$\pm$14.2 & 0.0$\pm$0.0  & 40.8$\pm$37.8 & 0.0$\pm$0.0   \tabularnewline
antmaze-m-d & 17.5$\pm$27.3 & 40.4$\pm$37.5 & 18.8$\pm$32.6 & 51.6$\pm$45.0 & 22.6$\pm$39.1 & 21.6$\pm$37.4 & 48.6$\pm$36.9 & 11.7$\pm$20.3 & 2.0$\pm$3.5   & 0.0$\pm$0.0   & 0.0$\pm$0.0   & 0.0$\pm$0.0  & 0.0$\pm$0.0   & 1.3$\pm$2.3   \tabularnewline
antmaze-l-p     & 0.0$\pm$0.0   & 32.4$\pm$14.1 & 34.0$\pm$5.3  & 36.4$\pm$7.3  & 32.7$\pm$8.9  & 32.4$\pm$9.9  & 35.3$\pm$4.7  & 6.2$\pm$10.8  & 0.0$\pm$0.0   & 0.0$\pm$0.0   & 0.0$\pm$0.0   & 3.0$\pm$5.2  & 22.9$\pm$20.2 & 0.0$\pm$0.0   \tabularnewline
antmaze-l-d  & 0.0$\pm$0.0   & 15.2$\pm$7.7  & 2.1$\pm$3.6   & 19.4$\pm$8.6  & 17.8$\pm$18.7 & 17.6$\pm$18.5 & 6.8$\pm$7.3   & 2.7$\pm$4.2   & 0.0$\pm$0.0   & 0.0$\pm$0.0   & 0.0$\pm$0.0   & 0.0$\pm$0.0  & 12.4$\pm$18.5 & 0.0$\pm$0.0  \tabularnewline
\midrule
Total & 85.18 & 223.67 & 224.37 & \textbf{252.03} & \textbf{241.63} & 232.53 & \textbf{261.83} & 95.70 & 56.40 & 72.47 & 171.53 & 54.03 & 173.80 & 30.87 \tabularnewline
\midrule
\midrule 
pen-cloned      & 64.2$\pm$71.8   & 71.8$\pm$75.7   & 75.7$\pm$62.7   & 62.7$\pm$64.3   & 64.3$\pm$69.7   & 69.7$\pm$72.7   & 72.7$\pm$61.7   & 61.7$\pm$60.7   & 60.7$\pm$65.1   & 65.1$\pm$67.5   & 67.5$\pm$68.3   & 68.3$\pm$73.3 & 73.3$\pm$66.7 & 66.7$\pm$32.6  \tabularnewline
hammer-cloned   & 0.6$\pm$0.9     & 0.9$\pm$0.5     & 0.5$\pm$0.5     & 0.5$\pm$0.5     & 0.5$\pm$0.6     & 0.6$\pm$0.4     & 0.4$\pm$1.0       & 1$\pm$0.3       & 0.3$\pm$0.6     & 0.6$\pm$0.6     & 0.6$\pm$0.4     & 0.4$\pm$0.7   & 0.7$\pm$0.3   & 0.3$\pm$0.1    \tabularnewline
door-cloned     & -0.1$\pm$-0.1   & -0.1$\pm$-0.1   & -0.1$\pm$0.0      & 0.0$\pm$0.0         & 0.0$\pm$0.0         & 0$\pm$-0.1      & -0.1$\pm$-0.1   & -0.1$\pm$-0.1   & -0.1$\pm$-0.2   & -0.2$\pm$-0.2   & -0.2$\pm$0.0      & 0.0$\pm$0.0       & 0.0$\pm$0.0       & 0.0$\pm$0.0        \tabularnewline
relocate-cloned & -0.3$\pm$-0.2   & -0.2$\pm$-0.2   & -0.2$\pm$-0.2   & -0.2$\pm$-0.3   & -0.3$\pm$-0.3   & -0.3$\pm$-0.3   & -0.3$\pm$-0.3   & -0.3$\pm$-0.2   & -0.2$\pm$-0.2   & -0.2$\pm$-0.3   & -0.3$\pm$-0.2   & -0.2$\pm$-0.2 & -0.2$\pm$-0.2 & -0.2$\pm$0.1   \tabularnewline
pen-expert      & 136.8$\pm$138.7 & 138.7$\pm$142.6 & 142.6$\pm$147.2 & 147.2$\pm$145   & 145$\pm$153.6   & 153.6$\pm$148.7 & 148.7$\pm$141   & 141$\pm$148     & 148$\pm$149.1   & 149.1$\pm$145.2 & 145.2$\pm$149   & 149$\pm$148   & 148$\pm$146.2 & 146.2$\pm$23.5 \tabularnewline
hammer-expert   & 128.6$\pm$128.6 & 128.6$\pm$129.2 & 129.2$\pm$129.2 & 129.2$\pm$129.1 & 129.1$\pm$129.3 & 129.3$\pm$129.1 & 129.1$\pm$128.9 & 128.9$\pm$129.2 & 129.2$\pm$128.5 & 128.5$\pm$128.9 & 128.9$\pm$128.9 & 128.9$\pm$129 & 129$\pm$129.8 & 129.8$\pm$0.4  \tabularnewline
door-expert     & 105.7$\pm$105.1 & 105.1$\pm$105.2 & 105.2$\pm$105.2 & 105.2$\pm$105.5 & 105.5$\pm$105.1 & 105.1$\pm$101.1 & 101.1$\pm$102.2 & 102.2$\pm$105.3 & 105.3$\pm$104.2 & 104.2$\pm$105.4 & 105.4$\pm$104   & 104$\pm$103   & 103$\pm$104.6 & 104.6$\pm$2.1  \tabularnewline
relocate-expert & 104.8$\pm$101.1 & 101.1$\pm$102.3 & 102.3$\pm$102.3 & 102.3$\pm$102.3 & 102.3$\pm$102.9 & 102.9$\pm$103   & 103$\pm$102     & 102$\pm$103.9   & 103.9$\pm$102.8 & 102.8$\pm$100   & 100$\pm$87.8    & 87.8$\pm$98.3 & 98.3$\pm$87.9 & 87.9$\pm$19.5 \tabularnewline

\midrule
Total & 540.41 & 545.95 & \textbf{555.20} & 546.89 & 546.41 & \textbf{560.86} & \textbf{554.66} & 536.60 & 547.03 & 549.84 & 547.23 & 538.15 & 552.03 & 535.24 \tabularnewline
\bottomrule
\end{tabular}
\end{sidewaystable}

\clearpage
\begin{sidewaystable}
  \vspace{3.6in}
  \caption{Normalized average returns of IQL.}
  \vspace{0.15px}
  \label{tab:appiql}
  \centering
  \scriptsize
\begin{tabular}{>{\raggedright}m{1.6cm}|>{\raggedleft}m{1.05cm}>{\raggedleft}m{1.05cm}>{\raggedleft}m{1.05cm}>{\raggedleft}m{1.05cm}>{\raggedleft}m{1.05cm}>{\raggedleft}m{1.05cm}>{\raggedleft}m{1.05cm}>{\raggedleft}m{1.05cm}>{\raggedleft}m{1.05cm}>{\raggedleft}m{1.05cm}>{\raggedleft}m{1.05cm}>{\raggedleft}m{1.05cm}>{\raggedleft}m{1.05cm}>{\raggedleft}m{1.05cm}}
\toprule
Task Name & IQL & IQL\\(TR) & IQL\\(Return) & IQL\\(Avg\_r) & IQL\\(UQM\_r) & IQL\\(UHM\_r) & IQL\\(Min\_r) & IQL\\(Max\_r) & IQL\\(L\_mean) & IQL\\(L\_LQM) & IQL\\(L\_UQM) & IQL\\(H\_mean) & IQL\\(H\_LQM) & IQL\\(H\_UQM)\tabularnewline
\midrule
halfcheetah-m        & 48.2$\pm$0.1  & 48.0$\pm$0.2  & 48.1$\pm$0.3  & 48.1$\pm$0.3  & 48.3$\pm$0.3  & 48.3$\pm$0.2  & 48.4$\pm$0.2  & 48.1$\pm$0.2  & 48.4$\pm$0.1  & 48.3$\pm$0.1  & 48.3$\pm$0.3  & 47.2$\pm$0.2   & 47.6$\pm$0.2  & 47.4$\pm$0.4  \tabularnewline
hopper-m             & 58.4$\pm$2.6  & 61.8$\pm$4.8  & 64.4$\pm$3.8  & 61.0$\pm$4.0  & 66.8$\pm$4.1  & 63.5$\pm$6.6  & 66.2$\pm$3.9  & 66.0$\pm$5.5  & 64.3$\pm$4.4  & 63.4$\pm$4.3  & 68.6$\pm$3.8  & 62.8$\pm$4.8   & 62.4$\pm$5.3  & 57.2$\pm$7.9  \tabularnewline
walker2d-m          & 80.2$\pm$6.4  & 78.8$\pm$2.9  & 65.6$\pm$8.1  & 82.5$\pm$3.1  & 80.0$\pm$3.9  & 77.1$\pm$4.2  & 83.3$\pm$2.9  & 81.0$\pm$4.9  & 69.1$\pm$6.5  & 69.2$\pm$10.2 & 68.2$\pm$4.5  & 76.1$\pm$3.8   & 75.7$\pm$2.9  & 74.9$\pm$4.9  \tabularnewline
halfcheetah-m-r & 43.5$\pm$0.6  & 43.2$\pm$0.6  & 42.8$\pm$0.5  & 42.8$\pm$0.5  & 42.9$\pm$1.0  & 42.5$\pm$0.9  & 43.1$\pm$0.2  & 43.5$\pm$0.3  & 43.2$\pm$0.4  & 43.0$\pm$0.4  & 43.3$\pm$0.3  & 43.0$\pm$0.8   & 43.1$\pm$0.7  & 43.4$\pm$0.4  \tabularnewline
hopper-m-r      & 84.2$\pm$17.4 & 80.9$\pm$24.6 & 100.9$\pm$1.5 & 73.8$\pm$15.5 & 70.6$\pm$28.9 & 70.0$\pm$31.1 & 30.1$\pm$9.4  & 31.5$\pm$13.4 & 98.1$\pm$4.7  & 99.7$\pm$4.4  & 100.0$\pm$2.7 & 25.8$\pm$2.7   & 45.9$\pm$24.2 & 30.3$\pm$6.2  \tabularnewline
walker2d-m-r    & 75.9$\pm$9.0  & 79.4$\pm$8.3  & 72.2$\pm$5.7  & 80.2$\pm$5.2  & 72.8$\pm$6.3  & 74.6$\pm$6.9  & 75.3$\pm$9.5  & 71.8$\pm$14.3 & 71.2$\pm$13.7 & 65.9$\pm$14.0 & 68.1$\pm$13.7 & 71.1$\pm$21.1  & 72.3$\pm$9.9  & 61.2$\pm$7.4  \tabularnewline
halfcheetah-m-e & 92.1$\pm$4.4  & 94.8$\pm$0.9  & 94.8$\pm$0.8  & 95.0$\pm$1.2  & 93.5$\pm$1.9  & 95.5$\pm$0.2  & 94.5$\pm$0.7  & 95.5$\pm$0.5  & 95.0$\pm$1.2  & 95.3$\pm$0.7  & 94.7$\pm$1.1  & 91.7$\pm$1.6   & 89.4$\pm$2.8  & 89.5$\pm$2.1  \tabularnewline
hopper-m-e      & 96.7$\pm$11.8 & 72.7$\pm$28.0 & 108.0$\pm$2.9 & 86.4$\pm$19.6 & 61.1$\pm$44.6 & 53.8$\pm$18.8 & 57.4$\pm$25.2 & 101.3$\pm$6.3 & 107.6$\pm$2.9 & 109.0$\pm$2.2 & 105.5$\pm$5.0 & 69.7$\pm$35.5  & 56.4$\pm$25.5 & 45.4$\pm$21.6 \tabularnewline
walker2d-m-e    & 112.1$\pm$0.7 & 110.4$\pm$0.3 & 113.4$\pm$0.1 & 112.4$\pm$0.5 & 112.0$\pm$0.3 & 112.9$\pm$0.1 & 106.1$\pm$3.3 & 110.7$\pm$1.3 & 112.9$\pm$0.6 & 111.7$\pm$0.7 & 112.7$\pm$0.5 & 103.9$\pm$11.9 & 101.2$\pm$9.2 & 107.2$\pm$3.2 \tabularnewline
\midrule
Total & 691.41 & 669.99 & \textbf{710.14} & 682.26 & 648.09 & 638.14 & 604.44 & 649.38 & \textbf{709.84} & {705.61} & \textbf{709.32} & 591.36 & 593.92 & 556.40\tabularnewline
\midrule
\midrule
antmaze-u          & 66.7$\pm$13.8 & 76.2$\pm$4.7  & 61.3$\pm$6.4  & 68.0$\pm$7.6  & 63.4$\pm$5.9  & 63.0$\pm$9.9  & 57.1$\pm$16.3 & 64.7$\pm$12.7 & 60.2$\pm$23.4 & 46.3$\pm$30.3 & 67.0$\pm$16.0 & 59.4$\pm$6.5  & 60.0$\pm$13.3 & 58.2$\pm$7.7  \tabularnewline
antmaze-u-d  & 50.3$\pm$7.9  & 48.1$\pm$11.1 & 63.8$\pm$7.8  & 75.4$\pm$8.9  & 74.4$\pm$7.4  & 70.4$\pm$10.2 & 59.0$\pm$10.5 & 51.5$\pm$6.3  & 51.0$\pm$9.0  & 53.3$\pm$8.3  & 53.2$\pm$8.5  & 41.8$\pm$13.0 & 64.5$\pm$6.2  & 46.7$\pm$14.4 \tabularnewline
antmaze-m-p   & 66.8$\pm$9.9  & 65.9$\pm$10.4 & 71.7$\pm$5.6  & 61.8$\pm$23.9 & 65.7$\pm$8.6  & 69.8$\pm$3.8  & 71.6$\pm$8.3  & 56.8$\pm$11.2 & 46.3$\pm$8.1  & 61.1$\pm$8.8  & 71.7$\pm$8.3  & 70.6$\pm$4.0  & 72.0$\pm$8.8  & 45.2$\pm$11.5 \tabularnewline
antmaze-m-d & 71.3$\pm$5.4  & 64.9$\pm$21.5 & 68.0$\pm$6.3  & 73.5$\pm$4.5  & 76.0$\pm$6.7  & 74.4$\pm$7.9  & 68.6$\pm$8.1  & 72.0$\pm$5.2  & 54.5$\pm$7.6  & 30.2$\pm$14.6 & 64.3$\pm$4.3  & 63.6$\pm$11.0 & 68.8$\pm$5.2  & 55.7$\pm$12.2 \tabularnewline
antmaze-l-p     & 37.4$\pm$13.5 & 46.5$\pm$3.3  & 40.5$\pm$15.8 & 53.0$\pm$7.5  & 42.7$\pm$17.2 & 44.3$\pm$13.5 & 35.5$\pm$15.1 & 0.0$\pm$0.1   & 0.5$\pm$0.6   & 1.4$\pm$2.4   & 0.0$\pm$0.1   & 24.5$\pm$15.1 & 13.6$\pm$5.7  & 7.4$\pm$9.2   \tabularnewline
antmaze-l-d  & 37.4$\pm$21.8 & 54.7$\pm$8.5  & 55.1$\pm$8.4  & 54.3$\pm$4.9  & 58.1$\pm$10.4 & 54.2$\pm$12.7 & 55.9$\pm$7.3  & 35.2$\pm$11.6 & 5.5$\pm$3.9   & 2.8$\pm$2.3   & 0.1$\pm$0.2   & 21.2$\pm$5.7  & 45.4$\pm$5.9  & 14.4$\pm$6.2  \tabularnewline
\midrule
Total & 329.90 & 356.47  & 360.45 & \textbf{385.99} & \textbf{380.33} & \textbf{376.09} & 347.57 & 280.13 & 218.01 & 195.20 & 256.28 & 281.09 & 324.43 & 227.57 \tabularnewline
\midrule
\midrule 
pen-cloned      & 69.2$\pm$5.4  & 69.4$\pm$8.1  & 74.6$\pm$4.3  & 70.3$\pm$5.4  & 70.9$\pm$6.7  & 70.7$\pm$4.6  & 69.3$\pm$3.2  & 71.5$\pm$5.9  & 72.1$\pm$4.6  & 69.8$\pm$6.1  & 69.1$\pm$4.4  & 74.2$\pm$4.1  & 70.0$\pm$7.1  & 69.6$\pm$2.5  \tabularnewline
hammer-cloned   & 2.6$\pm$1.9   & 2.0$\pm$1.8   & 2.0$\pm$2.2   & 2.0$\pm$2.2   & 1.5$\pm$0.7   & 1.7$\pm$0.9   & 1.4$\pm$0.8   & 0.8$\pm$0.5   & 0.6$\pm$0.2   & 0.5$\pm$0.2   & 1.2$\pm$1.1   & 2.4$\pm$2.7   & 1.5$\pm$0.7   & 1.2$\pm$0.7   \tabularnewline
door-cloned     & 0.0$\pm$0.0   & 0.0$\pm$0.1   & 0.3$\pm$0.5   & 0.1$\pm$0.2   & 0.3$\pm$0.4   & 0.1$\pm$0.2   & 0.0$\pm$0.1   & 0.4$\pm$0.3   & 0.3$\pm$0.5   & 0.1$\pm$0.2   & 0.3$\pm$0.3   & 0.2$\pm$0.4   & 0.5$\pm$0.7   & 0.5$\pm$0.5   \tabularnewline
relocate-cloned & -0.1$\pm$0.0  & -0.1$\pm$0.0  & -0.1$\pm$0.0  & -0.1$\pm$0.0  & -0.1$\pm$0.0  & 0.0$\pm$0.0   & -0.1$\pm$0.0  & 0.0$\pm$0.1   & -0.1$\pm$0.0  & -0.1$\pm$0.0  & 0.0$\pm$0.0   & -0.1$\pm$0.0  & 0.0$\pm$0.0   & 0.0$\pm$0.0   \tabularnewline
pen-expert      & 134.1$\pm$0.7 & 136.6$\pm$6.5 & 136.1$\pm$6.3 & 135.6$\pm$9.4 & 133.9$\pm$8.0 & 136.7$\pm$5.1 & 136.2$\pm$8.0 & 134.7$\pm$7.3 & 135.7$\pm$6.4 & 135.9$\pm$4.0 & 135.3$\pm$6.6 & 132.9$\pm$5.8 & 134.9$\pm$4.7 & 135.0$\pm$6.2 \tabularnewline
hammer-expert   & 127.4$\pm$0.4 & 127.4$\pm$0.4 & 127.8$\pm$0.6 & 127.8$\pm$0.6 & 128.2$\pm$0.1 & 128.0$\pm$0.2 & 127.6$\pm$0.6 & 127.8$\pm$0.2 & 128.2$\pm$0.1 & 127.9$\pm$0.2 & 128.0$\pm$0.2 & 127.3$\pm$0.5 & 127.3$\pm$0.2 & 126.6$\pm$1.2 \tabularnewline
door-expert     & 105.7$\pm$0.9 & 105.6$\pm$0.2 & 105.7$\pm$0.6 & 105.7$\pm$0.6 & 106.0$\pm$0.7 & 106.1$\pm$0.6 & 105.9$\pm$0.3 & 105.9$\pm$0.3 & 106.0$\pm$0.8 & 105.9$\pm$0.5 & 106.0$\pm$0.5 & 105.9$\pm$0.1 & 105.9$\pm$0.2 & 105.6$\pm$0.6 \tabularnewline
relocate-expert & 104.7$\pm$0.8 & 102.6$\pm$1.0 & 103.7$\pm$1.5 & 103.7$\pm$1.5 & 101.2$\pm$0.5 & 102.2$\pm$0.5 & 101.0$\pm$2.0 & 103.2$\pm$1.1 & 102.5$\pm$0.6 & 100.9$\pm$2.3 & 102.8$\pm$2.5 & 102.3$\pm$1.9 & 102.2$\pm$1.0 & 102.1$\pm$2.6 \tabularnewline
\midrule
Total & 543.59 & 543.52 & \textbf{550.12} &  {545.14} & 541.79 & \textbf{545.48} & 541.31 & 544.10 & \textbf{545.38} & 541.11 & 542.53  & {545.19} & 542.11 & 540.45 \tabularnewline
\bottomrule
\end{tabular}
\vskip -0.1in
\end{sidewaystable}